\documentclass[10pt,journal,compsoc]{IEEEtran}

\ifCLASSOPTIONcompsoc
  \usepackage[nocompress]{cite}
\else
  \usepackage{cite}
\fi

\ifCLASSINFOpdf
\else
\fi

\usepackage{ragged2e} 
\usepackage[permil]{overpic}
\usepackage{arydshln}  
\usepackage{amsmath}
\usepackage{amssymb}
\usepackage{amsfonts}
\usepackage{booktabs} 
\usepackage{multirow}
\usepackage{algorithm}
\usepackage{algorithmic}
\usepackage{xcolor}
\usepackage{mathtools}
\usepackage{amsthm}
\usepackage{array}
\usepackage[caption=false,font=normalsize,labelfont=sf,textfont=sf]{subfig}
\usepackage{textcomp}
\usepackage{stfloats}
\usepackage{url}
\usepackage{verbatim}
\usepackage{graphicx}
\usepackage{bbding}
\usepackage{rotating}
\usepackage{color}
\usepackage[capitalize,noabbrev]{cleveref}
\usepackage[permil]{overpic}
\usepackage[table]{xcolor}
\usepackage{newtxtext}  

\newcommand\et{\textit{et al.}} 
\newcommand\gray{\cellcolor{gray!20}} 
\newcommand{\ours}[0]{\textsc{CapNet}}
\definecolor{amber}{rgb}{1.0, 0.75, 0.0}
\definecolor{darkcerulean}{rgb}{0.03, 0.27, 0.49}
\definecolor{darkcoral}{rgb}{0.8, 0.36, 0.27}

\begin{document}
\title{Unleashing the Power of Vision-Language Models for Long-Tailed Multi-Label Visual Recognition}

\author{
Wei Tang, Zuo-Zheng Wang, Kun Zhang, Tong Wei, and~Min-Ling~Zhang, Senior Member, IEEE
\IEEEcompsocitemizethanks{
\IEEEcompsocthanksitem Wei Tang is with the School of Computer Science and Engineering, Southeast University, Nanjing 210096, China, the Key Laboratory of Computer Network and Information Integration (Southeast University), MoE, China, and with the Mohamed bin Zayed University of Artificial Intelligence (MBZUAI), Abu Dhabi, UAE (e-mail: tangw@seu.edu.cn). \protect\\
\IEEEcompsocthanksitem Zuo-Zheng Wang, Tong Wei and Min-Ling Zhang are with the School of Computer Science and Engineering, Southeast University, Nanjing 210096, China, and the Key Laboratory of Computer Network and Information Integration (Southeast University), MoE, China (e-mail: wendy813wzz@126.com; weit@seu.edu.cn; zhangml@seu.edu.cn).\protect\\
\IEEEcompsocthanksitem Kun Zhang is with the Carnegie Mellon University, Pittsburgh, PA 15213 USA, and with the Mohamed bin Zayed University of Artificial Intelligence (MBZUAI), Abu Dhabi, UAE (e-mail: kunz1@cmu.edu). \protect\\
}
}

\IEEEtitleabstractindextext{
\begin{abstract}
\justifying
Long-tailed multi-label visual recognition poses a significant challenge, as images typically contain multiple labels with highly imbalanced class distributions, leading to biased models that favor head classes while underperforming on tail classes. Recent efforts have leveraged pre-trained vision-language models, such as CLIP, alongside long-tailed learning techniques to exploit rich visual-textual priors for improved performance. However, existing methods often derive semantic inter-class relationships directly from imbalanced datasets, resulting in unreliable correlations for tail classes due to data scarcity. Moreover, CLIP's zero-shot paradigm is optimized for single-label image-text matching, making it suboptimal for multi-label tasks. To address these issues, we propose the correlation adaptation prompt network ({\ours}), a novel end-to-end framework that explicitly models label correlations from CLIP's textual encoder. The framework incorporates a graph convolutional network for label-aware propagation and learnable soft prompts for refined embeddings. It utilizes a distribution-balanced Focal loss with class-aware re-weighting for optimized training under imbalance. Moreover, it improves generalization through test-time ensembling and realigns visual-textual modalities using parameter-efficient fine-tuning to avert overfitting on tail classes without compromising head class performance. 
Extensive experiments and ablation studies on benchmarks including VOC-LT, COCO-LT, and NUS-WIDE demonstrate that {\ours} achieves substantial improvements over state-of-the-art methods, validating its effectiveness for real-world long-tailed multi-label visual recognition.
\end{abstract}

}

\maketitle

\IEEEdisplaynontitleabstractindextext

\IEEEpeerreviewmaketitle

\IEEEraisesectionheading{\section{Introduction}\label{sec:intro}}
\IEEEPARstart{L}{ong-tailed} visual recognition is a challenging area associated with learning from highly imbalanced data, where head classes dominate with abundant samples while tail classes suffer from scarcity~\cite{ZhangKHYF23, yun2019cutmix, alshammari2022long, menon2021long}. Recent progress in long-tailed visual recognition has mainly centered on the single-label multi-class setting. However, real-world images often encompass multiple objects and concepts, giving rise to long-tailed multi-label visual recognition (LTML), where each image is associated with multiple labels exhibiting long-tailed distributions \cite{wu2020distribution, guo2021long, chen2023class, xia2023lmpt, Yan0MH024, TimmermannJKL25}. LTML introduces additional intricacies compared to its single-label multi-class counterpart, including label co-occurrences, which create interdependencies among classes, and intra-class imbalances between positive and negative instances.

\begin{figure}[!t]
\setlength{\abovecaptionskip}{0.cm} 
    \centering
  \begin{overpic}[width=8.5cm]{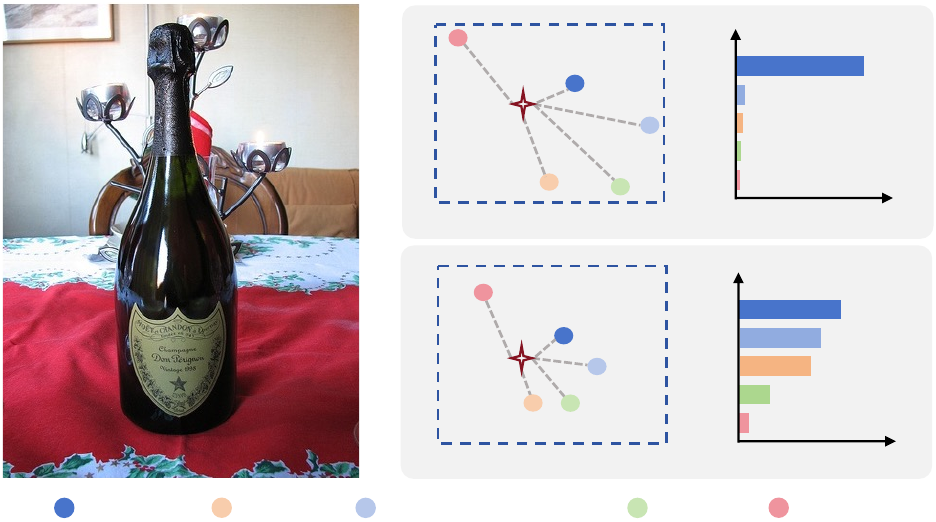}
        \put(570, 309){\footnotesize (a) zero-shot CLIP}
        \put(800, 510){\tiny $p(\boldsymbol{y}|\boldsymbol{x})$}
        \put(930, 480){\tiny 0.97}
        \put(802, 450){\tiny 0.02}
        \put(800, 420){\tiny 0.0004}
        \put(800, 389){\tiny 0.0007}
        \put(800, 358){\tiny 0.00008}   
        \put(650, 53){\footnotesize (b) {\ours}}
        \put(800, 250){\tiny $p(\boldsymbol{y}|\boldsymbol{x})$}
        \put(905, 220){\tiny 0.78}
        \put(885, 190){\tiny 0.56}
        \put(875, 160){\tiny 0.54}
        \put(830, 129){\tiny 0.21}
        \put(808, 98){\tiny 0.02}
        \put(85, 5){\small bottle}
        \put(256, 5){\small chair}
        \put(410, 5){\small dinner table}
         \put(700, 5){\small sofa}
          \put(850, 5){\small sheep}
    \end{overpic}
    \caption{Comparison between zero-shot CLIP~\cite{radford2021learning} and our proposed {\ours}. Image embeddings are represented by red four-angle stars, and text embeddings by circles. The proximity indicates the higher probability that the category of text embeddings matches the image. 
}
 \label{fig:capn_illu}
\end{figure}
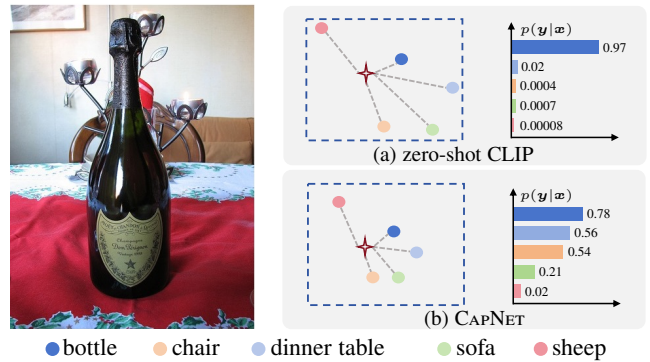

Existing LTML methods have primarily relied on strategies like class re-sampling, loss re-weighting, and specialized architectures, typically built upon ImageNet-pre-trained convolutional neural networks (CNNs) as backbones~\cite{wu2020distribution, guo2021long, chen2023class}. Motivated by the emergence of vision-language models (VLMs) like contrastive language-image pre-training (CLIP)~\cite{radford2021learning}, recent works such as LMPT~\cite{xia2023lmpt} have pioneered the use of CLIP for long-tailed multi-label visual recognition, which introduces additional captions and class-specific embedding losses to learn fine-grained and class-related embeddings.

Despite these advances, CLIP's zero-shot paradigm is inherently designed for single-label image-text matching, making it suboptimal for multi-label tasks. As illustrated in Fig.~\ref{fig:capn_illu}, zero-shot CLIP tends to prioritize the most salient class, e.g., assigning $0.97$ confidence to \emph{bottle}, neglecting co-occurring labels such as \emph{dinner table}. Intuitively, leveraging label co-occurrences is a feasible way to mitigate long-tailed imbalances, as it enables knowledge transfer from head to tail classes. For example, the tail-class \emph{parking meters} often accompany the head-class \emph{cars}, while the medium-class \emph{keyboards} are commonly found with the head-class \emph{laptops}. Consequently, such label co-occurrences can synchronously enhance the performance of head-to-tail categories. While LMPT~\cite{xia2023lmpt} attempts to model these dependencies using shared image-caption pairs, it suffers from noisy and incomplete semantic representations: (1) the captions are generated by models or annotated manually, introducing noise and high costs; (2) the captions inherit the long-tailed biases of the datasets, failing to capture nuanced inter-class connections, particularly for tail classes. Therefore, the model may not capture all the nuanced connections between various classes, which potentially limits the effectiveness in the long-tailed multi-label visual recognition scenarios. \looseness=-1

To address these limitations and unleash the potential of VLMs for LTML, we propose the correlation adaptation prompt network ({\ours}), an end-to-end framework that adapts CLIP for robust LTML. Firstly, {\ours} enables the computation of independent probabilities for each label, thereby achieving genuine multi-label compatibility without mutual exclusivity assumptions. Secondly, {\ours} explicitly captures label correlations via a graph convolutional network (GCN)~\cite{chen2019multi, ding2023exploring}, constructing a robust correlation matrix from CLIP's textual encoder rather than noisy captions-derived priors. 
Thirdly, we adapt the distribution-balanced Focal loss~\cite{wu2020distribution} with re-balanced weights tailored for class-aware sampling~\cite{shen2016relay}. Finally, we introduce a test-time ensembling strategy that aggregates diverse predictions to enhance generalization, and a parameter-efficient fine-tuning (PEFT) method, which learns a sparse set of task-specific parameters to prevent overfitting, especially on tail classes. 

To verify the effectiveness of our proposed {\ours}, extensive evaluations on benchmarks like VOC-LT, COCO-LT, and NUS-WIDE~\cite{wu2020distribution, everingham2015pascal, lin2014microsoft} demonstrate that {\ours} achieves substantial performance gains over state-of-the-art approaches, highlighting its practical utility. Our contributions can be summarized as follows:     \looseness=-1
\begin{itemize}
\item We propose {\ours}, an end-to-end framework that adapts CLIP for LTML, incorporating independent label predictions, a GCN-based module for robust label correlations, and a distribution-balanced Focal loss with class-aware re-balancing. \looseness=-1
\item We introduce a test-time ensembling strategy and a parameter-efficient fine-tuning approach with sparse task-specific parameters to enhance generalization and mitigate overfitting, especially for tail classes.
\item Through comprehensive experiments on benchmarks such as VOC-LT, COCO-LT, and NUS-WIDE, we demonstrate that {\ours} significantly outperforms state-of-the-art methods, validating its effectiveness.
\end{itemize}

The remainder of this paper is organized as follows: Section~\ref{sec:related} reviews related work. Section~\ref{sec:methodology} elucidates the preliminaries and {\ours}'s technical details. Section~\ref{sec:experiments} reports experimental results, and Section~\ref{sec:con} concludes the paper.

\section{Related Work}\label{sec:related}
\subsection{Multi-Label Visual Recognition}
Multi-label visual recognition presents a core challenge in machine learning~\cite{zhang2025la, wen2025learning}. In contrast to multi-class recognition, where each instance receives a single label, multi-label classification assigns multiple labels to an instance. Early approaches relied on independent binary classifiers~\cite{zhang2013review, tsoumakas2007multi} or k-nearest neighbors~\cite{zhang2007ml}, often overlooking label correlations. Research has demonstrated that exploiting these correlations improves model generalization. To this end, Wang \et~\cite{wang2016cnn} integrated convolutional and recurrent neural networks to model label dependencies explicitly. Subsequent methods adopted graph neural networks for embedding such correlations~\cite{chen2019multi, chen2019learning, wang2020multi, ye2020attention}. Nevertheless, these approaches heavily depend on the quality of label supervision.

To alleviate annotation burdens, multi-label learning with partial labels emerged, providing only label subsets. Huynh \et~\cite{huynh2020interactive} inferred missing labels via mini-batch co-occurrences, while Chen \et~\cite{chen2022structured} incorporated semantic correlations. These techniques, however, fall short in imbalanced distributions. Recent advancements leverage vision-language models to align visual and textual features in multi-label tasks~\cite{sun2022dualcoop, guo2023texts, ding2023exploring, xia2023lmpt, Yan0MH024, TimmermannJKL25}. Despite these gains, existing methods neglect long-tailed distributions and underexploit label dependencies inherent in vision-language models. \looseness=-1

\subsection{Long-Tailed Visual Recognition}
Real-world datasets frequently follow long-tailed distributions, with head classes dominating sample counts while tail classes remain scarce~\cite{ZhangKHYF23, fang2023revisiting, he2025partialclip, WeiL20}. This imbalance causes models to favor head classes, degrading tail class accuracy. Balancing the learning process thus becomes essential. Simple data augmentation techniques address this by oversampling minorities~\cite{chawla2002smote, chris2003class}. Balanced feature learning further mitigates biases~\cite{tang2023demipl, kang2019decoupling, cui2021parametric, li2022targeted, tang2023miplgp}. Recent optimization strategies prioritize tail classes~\cite{lin2020focal, park2021influence, cao2019learning}, while inference-time adjustments enhance robustness~\cite{cai2021ace, wang2020long}. Incorporating textual modalities has also shown potential for refining visual-language representations in long-tailed visual recognition tasks~\cite{ma2021simple, tian2022vl}.

For long-tailed multi-label recognition, Wu \et~\cite{wu2020distribution} devised a distribution-balanced loss by extending re-balanced sampling and cost-sensitive re-weighting, though it sometimes compromises head class performance. To harmonize sampling strategies, Guo \et~\cite{guo2021long} drew from bilateral-branch networks~\cite{zhou2020bbn} for long-tailed recognition, merging re-balanced and uniform approaches. Advancing interaction modeling, Chen \et~\cite{chen2023class} proposed the class-aware embedding network with multi-head attention on visual features and GloVe-derived label embeddings~\cite{pennington2014glove}. Extending to imperfect settings with partial labels, Zhang \et~\cite{ZhangLZOTZ23} developed a unified framework for joint handling of long-tailed and partial-label challenges. For greater robustness to distribution shifts, Lin \et~\cite{LinPCXQC24} introduced a distributionally robust loss optimizing worst-case outcomes. Prompt-based methods build on this by adapting CLIP via tuning and textual cues~\cite{xia2023lmpt}. Refining these prompts category-wise, Yan \et~\cite{Yan0MH024} tailored feature learning for enhanced alignment in long-tailed multi-label scenarios. Finally, Tao \et~\cite{TaoLWZCLHC25} explored neural collapse phenomena, proposing collapse-alleviating mechanisms to boost generalization.

\begin{figure*}[!t]
    \centering
  \begin{overpic}[width=17cm]{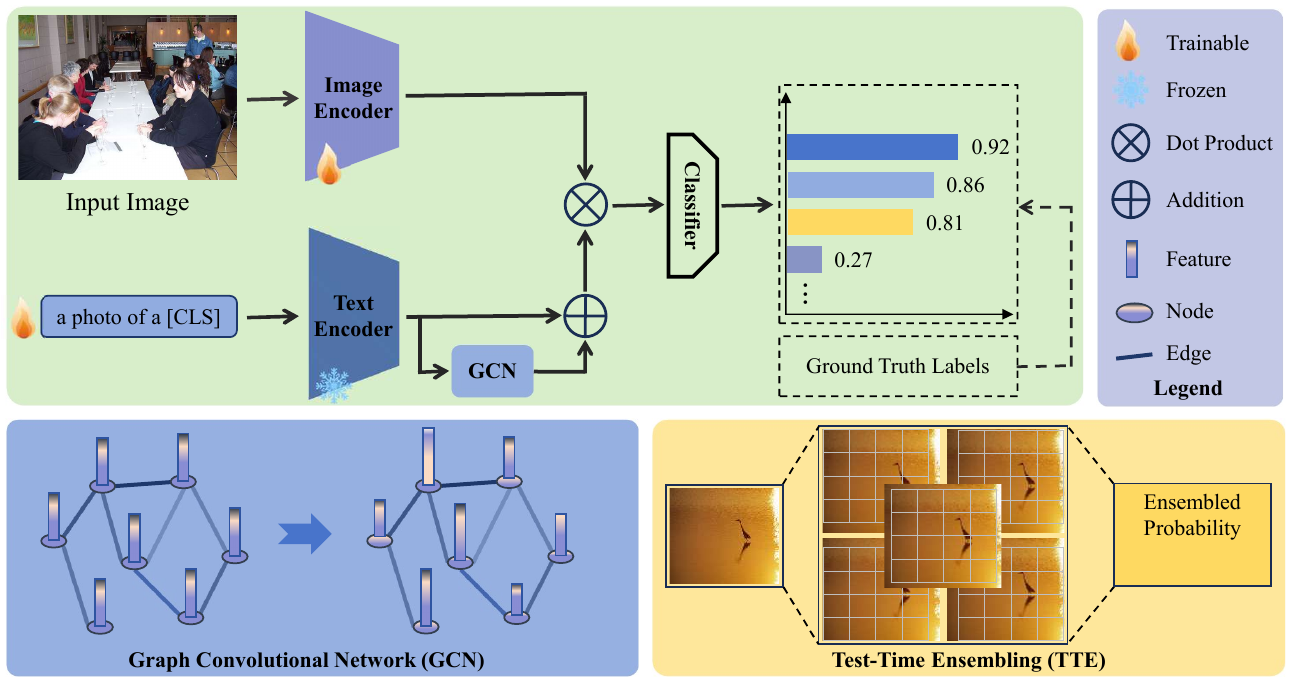}      
    \put(620, 435){\small $p(\boldsymbol{y}|\boldsymbol{x})$}
            \put(795, 380){\small $\mathcal{L}_{cls}$}
            \put(867, 88){\scriptsize $\boldsymbol{p}_{*} = \frac{1}{5} \sum_{cr} \boldsymbol{p}^{cr}$} 
    \end{overpic}
    \caption{Overall framework of our {\ours} method for long-tailed multi-label visual recognition. The image encoder is fine-tuned using either full fine-tuning or parameter-efficient fine-tuning.}
 \label{fig:capn_framework}
\end{figure*}

\subsection{Parameter-Efficient Fine-tuning Technique}
To transfer pre-trained models to downstream tasks, a directly adopted method is full fine-tuning, where all model parameters are updated. However, full fine-tuning often incurs computational overheads and is prone to severe over-fitting on long-tailed datasets due to limited tail-class samples~\cite{shi2023parameter, shi2025lift, zhangfx2025peft}. Recent methodologies have strategically adopted prompt tuning within the textual modality \cite{sun2022dualcoop, zhou2022conditional, xia2023lmpt}. This parameter-efficient technique has proven crucial in improving performance and addressing the aforementioned challenges.

To enhance the fine-tuning efficiency of current vision Transformer (ViT) models \cite{dosovitskiy2021image}, Jia \et~\cite{jia2022visual} introduced visual prompt tuning (VPT), which involves prepending learnable prompts to the patch embeddings and freezing the entire pre-trained backbone. Two variations of VPT exist: VPT-Shallow, which prepends prompts only at the first layer, and VPT-Deep, which prepends prompts at all layers. In addition to prompt-related approaches, Chen \et~\cite{chen2022adaptformer} proposed AdaptFormer to enhance ViT's transferability with minimal computational cost. AdaptFormer simplifies the process by replacing the original multi-layer perceptrons (MLP) block with AdaptMLP, featuring frozen branches and trainable bottlenecks to boost transferability at low cost.

\section{Methodology}
\label{sec:methodology}
In this section, we delineate the proposed {\ours} for addressing long-tailed multi-label visual recognition. The overall framework is illustrated in Fig.~\ref{fig:capn_framework}.

\par
\subsection{Preliminaries}
Let $\mathcal{X} \subseteq \mathbb{R}^{H \times W \times 3}$ denote the input image space, and $\mathcal{Y} = \{1, 2, \dots, C\}$ represent the label space comprising $C$ distinct classes. The goal of long-tailed multi-label learning is to learn a classifier $h: \mathcal{X} \to 2^{\mathcal{Y}}$ that accurately predicts the subset of relevant labels for each input image, despite severe class imbalances in the training data.
Consider a long-tailed multi-label training dataset $\mathcal{D}_{\text{train}} = {(\boldsymbol{x}^k, \boldsymbol{y}^k) \mid 1 \leq k \leq N }$, where $N$ is the number of samples, $\boldsymbol{x}^k \in \mathcal{X}$ is the $k$-th input image, and $\boldsymbol{y}^k = [y_1^k, y_2^k, \dots, y_C^k]^\top \in \{0,1\}^C$ is its corresponding binary label vector, with $y_c^k = 1$ indicating the presence of class $c$ in image $\boldsymbol{x}^k$. During inference, for a test image $\boldsymbol{x} \in \mathcal{D}_{\text{test}}$, {\ours} predicts the label set $\hat{\boldsymbol{y}} = h(\boldsymbol{x})$. 

To ground our approach, we briefly review the contrastive language-image pre-training (CLIP) model~\cite{radford2021learning}, which serves as the foundational backbone for {\ours}. CLIP comprises two encoders: a vision encoder, e.g., ResNet~\cite{he2016deep} or Vision Transformer~\cite{dosovitskiy2021image}, that maps images to embeddings, and a text encoder based on Transformer~\cite{VaswaniSPUJGKP17} that processes textual inputs. Pre-trained on 400 million image-text pairs using contrastive learning, CLIP aligns visual and textual representations in a shared embedding space, enabling zero-shot transfer to diverse downstream tasks. For classification, text prompts such as ``$\mathrm{a~photo~of~a ~\left[CLS\right]}$" are constructed for each class, and predictions are derived by computing cosine similarities between the image embedding and the text embeddings, typically followed by softmax normalization to yield class probabilities. This multimodal alignment is particularly advantageous for long-tailed multi-label visual recognition tasks, as it provides rich semantic priors that can mitigate data scarcity in tail classes by leveraging language-based knowledge transfer.

\subsection{Approach Overview}
We introduce the correlation adaptation prompt network ({\ours}), a novel framework designed to harness the capabilities of pre-trained VLMs for long-tailed multi-label visual recognition. Building upon CLIP~\cite{radford2021learning}, which comprises an image encoder $E_I(\cdot)$ and a text encoder $E_T(\cdot)$, {\ours} adapts these components to model label correlations and mitigate imbalances. Following the conditional prompt learning~\cite{zhou2022conditional}, we construct class-specific soft prompts as follows:
\begin{equation}
\boldsymbol{t}_{c}|_1^{M}=\left[\mathrm{V}\right]_{1}\left[\mathrm{V}\right]_{2}\ldots\left[\mathrm{V}\right]_{m}\ldots\left[\mathrm{V}\right]_{M}\left[\mathrm{CLS}\right],
\label{eq:init_tp}
\end{equation}
where $c \in \{1, \dots, C\}$, each $[\mathrm{V}]_m$ ($m \in {1, \dots, M}$) denotes a learnable context token with the same dimensionality as standard word embeddings, e.g., 512 for CLIP, and $[\mathrm{CLS}]$ is substituted with the actual class name, e.g., \emph{person}, \emph{horse}, or \emph{car}. Here, $M$ controls the prompt length.

During training, the text encoder processes these soft prompts to yield initial text embeddings. These embeddings are then input to a three-layer GCN to explicitly capture label correlations. The GCN employs a correlation matrix $\mathcal{G} \in \mathbb{R}^{C \times C}$, derived from CLIP's textual priors, to propagate semantic information across classes, producing refined outputs $\boldsymbol{H}_L$. As shown in Fig.~\ref{fig:capn_framework}, a residual connection integrates the original text features $\boldsymbol{F}_t^* = \boldsymbol{F}_t + \boldsymbol{H}_L = [\boldsymbol{f}_1^*, \boldsymbol{f}_2^*, \dots, \boldsymbol{f}_C^*] \in \mathbb{R}^{C \times d}$ at epoch $t$. For an input image $\boldsymbol{x}^k$, the image encoder extracts the visual embedding $\boldsymbol{v}^k = E_I(\boldsymbol{x}^k)$. The prediction probability for each class $c$ is then computed as follows:
\begin{equation}
      p(y_{c}^k \mid \boldsymbol{x}^k) = \sigma \left( \text{sim} (\boldsymbol{E_I}(\boldsymbol{x}^k),\boldsymbol{f}_{c}^*)/\tau \right),
  \label{eq:sampro}
\end{equation}
where $\sigma(\cdot)$ denotes the sigmoid function, $\operatorname{sim}(\cdot, \cdot)$ represents the cosine similarity, and $\tau$ is the learned temperature parameter for CLIP. This formulation supports independent multi-label predictions, differing from CLIP's softmax, designed for single-label classification. \looseness=-1

\subsection{GCN for Semantic Correlation Learning}
The GCN is tailored for processing graph-structured data, iteratively refining node representations by aggregating features from neighboring nodes to capture complex interdependencies within the graph.
Prior multi-label classification techniques, such as ML-GCN~\cite{chen2019multi}, have constructed label correlation matrices to propagate information across GCN nodes. Typically, these correlations are derived via conditional probabilities, which require balanced label distributions to yield reliable co-occurrence statistics. However, in long-tailed datasets, such approaches falter due to inherent imbalances, leading to unreliable estimates, particularly for tail classes where data scarcity amplifies estimation errors.

To circumvent these issues, we leverage CLIP's textual encoder to derive robust semantic priors. Specifically, we input class-specific prompts of the form ``$\mathrm{a~photo~of~a ~\left[CLS\right]}$" into the text encoder, obtaining embeddings $\boldsymbol{z}_c \in \mathbb{R}^d$ for each class $c \in \{1, \dots, C\}$. The correlation matrix is then computed as the pairwise cosine similarities:
\begin{equation}
  \mathbf{\mathcal{A}} = sim(\mathbf{\mathcal{Z}}^T, \mathbf{\mathcal{Z}}),
  \label{eq:co_m0}
\end{equation}
where $\mathcal{A}=(a_{ij})_{C\times C}$ and $\mathcal{Z} = [\boldsymbol{z}_1, \boldsymbol{z}_2, \dots, \boldsymbol{z}_C] \in \mathbb{R}^{d \times C}$.
To mitigate over-smoothing in graph propagation, we adjust the matrix to balance self-connections and neighbor influences:
\begin{equation}
  a^{\prime}_{ij}= \left\{ \begin{array}{l}
    ( s/\sum\nolimits_{i \ne j}^C a_{ij} ) \times a_{ij}, \quad\, \text{if~} i \ne j,\\
    1 - s,\qquad\qquad\qquad\quad \, \text{if~} i = j,
  \end{array} \right.
  \label{eq:co_m1}
\end{equation}
where $s \in [0,1]$ is a hyperparameter governing the trade-off between a node's own features and those of its neighbors.
The final adjacency matrix for the label correlation graph $\mathcal{G}$ is obtained by row-wise softmax normalization:
\begin{equation}
a_{ij}^{*}=\frac{\exp (a^{\prime}_{ij}/\tau^{\prime})} {\sum\nolimits_{j}^C \exp (a^{\prime}_{ij}/\tau^{\prime})},
\label{eq:co_m2}
\end{equation}
where $\tau'$ serves as a temperature parameter that controls the softness of the distribution. The final pairwise cosine similarity matrix is denoted as $\mathcal{A}^* = (a^*_{ij})_{C \times C}$.

This CLIP-derived strategy diverges from data-dependent methods by reducing vulnerability to long-tailed biases and enhancing correlation reliability through language-based priors. By integrating the semantics with soft prompts in {\ours}, we enable effective knowledge transfer from head to tail classes, fostering improved generalization in long-tailed multi-label visual recognition tasks and aligning with recent advances in multimodal graph learning for imbalanced domains.

\subsection{Long-Tailed Multi-Label Classification Loss}
The distribution-balanced Focal loss~\cite{wu2020distribution} incorporates re-balanced weights optimized for multi-class settings with class-aware sampling, which can lead to suboptimal handling of multi-label imbalances where positive and negative instances coexist variably across classes, potentially exacerbating overfitting on head classes and underfitting on tails.
To alleviate these challenges in long-tailed multi-label visual recognition, we adapt the re-balanced weights to mitigate multi-label imbalance in the following manner:
\begin{equation}
 r = \alpha + \sigma\left(\beta \times \left(\frac{1}{n_c / N}-\theta \right) \right),   
 \label{eq:re_weight}
\end{equation}
where $n_c = \sum_{i=1}^N y_c^i$ is the number of training samples containing class $c$, $N$ is the total number of samples, and $\sigma(\cdot)$ is the sigmoid function. The parameter $\alpha$ serves as a global scaling factor on the weight, while $\beta$ and $\theta$ determine the specific form of the mapping function. This formulation dynamically adjusts weights inversely with class frequency, ensuring tail classes receive amplified emphasis during optimization. The final long-tailed multi-label classification loss functions for each image and the whole dataset are formulated as follows: \looseness=-1
\begin{equation}
\begin{aligned}
{\ell}_{cls} & =  
            \begin{cases}
            -r\left(1-q_c^i\right)^\gamma\log\left(q_c^i\right)  & \text{if~}y_c^i = 1,\\
            -\frac{r}{\zeta}\left(q_c^i\right)^\gamma\log\left(1-q_c^i\right) & \text{if~} y_c^i = 0,
            \end{cases} \\
\end{aligned}
\label{eq:cls_loss}
\end{equation}
\begin{equation}
    \mathcal{L}_{cls} = \frac{1}{N} \sum_{i=1}^{N} {\ell}_{cls}.
    \label{eq:full_cls_loss}
\end{equation}
Here, $\gamma > 0$ is the focusing parameter that downweights easy examples. $q_c^i$ represents the modulated probability: for positive instances ($y_c^i = 1$), $q_c^i = \sigma(z_c^i - v_c)$; for negatives ($y_c^i = 0$), $q_{i c} = \sigma(\zeta (z_c^i - v_c))$. The class-specific bias $v_c$ introduces a frequency-dependent margin:
\begin{equation}
    v_c = \kappa \times \log \left(\frac{1}{n_c / N} - 1\right).
    \label{eq:class_bias}
\end{equation}

Here, hyperparameters $\kappa$ and $\zeta$ control the balancing and focusing mechanisms. This design not only mitigates the dominance of negative samples in multi-label scenarios but also synergizes with {\ours}'s VLM priors to enhance tail-class recall, proving essential for applications like fine-grained scene parsing where rare co-occurrences prevail.

\subsection{Test-Time Ensembling for {\ours}}
Vision Transformers (ViTs) partition input images into fixed-size patches, which may inadvertently split critical semantic patterns across patch boundaries, potentially degrading representation quality. To enhance generalization, particularly in long-tailed multi-label scenarios where tail classes benefit from diversified predictions, we adopt a test-time ensembling strategy. This involves aggregating logits from multiple perturbed versions of the test image to mitigate patch-induced artifacts and boost robustness.
Formally, for a test image $\boldsymbol{x}_{*}$, the ensembled probability $\boldsymbol{p}_{*} \in \mathbb{R}^C$ is computed as: \looseness=-1
\begin{equation}
    \boldsymbol{p}_{*} = \frac{1}{N_{\text{aug}}} \sum_{i=1}^{N_{\text{aug}}} \Omega_{i}(\boldsymbol{x}_{*}),
\end{equation}
where $\Omega_{i}(\cdot)$ denotes the model's probability output for the $i$-th augmented variant of $\boldsymbol{x}_{*}$, and $N_{\text{aug}}$ is the number of augmentations.
In practice, $\boldsymbol{x}_{*}$ is first resized to $(224 + e) \times (224 + e)$, followed by diverse 224$\times$224 croppings to generate the variants, with the final prediction derived from averaging the logits. Critically, $e$ must not be a multiple of the ViT patch size (typically 16) to ensure distinct patch compositions across crops; otherwise, substantial overlap diminishes augmentation diversity. By default, we set $e=24$ and $N_{\text{aug}}=5$, corresponding to center, top-left, top-right, bottom-left, and bottom-right crops. This method extends effectively to CNN-based backbones like ResNet-50, offering a lightweight yet potent means to elevate performance in imbalanced domains. \looseness=-1

\subsection{Parameter-Efficient Fine-Tuning for {\ours}}
Full fine-tuning harnesses downstream datasets to refine pre-trained models, yielding substantial performance gains. However, updating all parameters incurs high computational costs and risks severe overfitting, particularly on long-tailed datasets where tail classes suffer from data paucity~\cite{shi2023parameter, shi2025lift}. 
To overcome these limitations while preserving efficacy, we integrate AdaptFormer~\cite{chen2022adaptformer}, a leading parameter-efficient fine-tuning method renowned for its balance of efficiency and performance. AdaptFormer augments the original MLP block with a parallel bottleneck sub-branch, comprising a down-projection $W_{\text{down}} \in \mathbb{R}^{d^\prime \times \hat{d}}$ and an up-projection $W_{\text{up}} \in \mathbb{R}^{\hat{d} \times d^\prime}$, where $\hat{d} \ll d^\prime$ denotes the reduced intermediate dimension. The output of the $\ell$-th encoder block is formulated as:
\begin{equation}
\begin{split}
h_{\ell} = s \cdot \text{ReLU}(\text{LN}(h_{\ell}^{\prime}) \cdot W_{\text{down}}) \cdot W_{\text{up}}  
+ \text{MLP}(\text{LN}(h_{\ell}^{\prime})) + h_{\ell}^{\prime},
\end{split}
\end{equation}
where $h_{\ell}^{\prime}$ represents the output from the $\ell$-th multi-head self-attention layer~\cite{VaswaniSPUJGKP17}, $\operatorname{LN}(\cdot)$ signifies layer normalization, and $s$ is a learnable scaling factor.
This architecture maintains the generalization of image-text pre-training. By introducing only a small set of task-specific parameters, the model not only mitigates overfitting but also achieves rapid convergence. It is worth noting that the PEFT technique is tailored for the ViT-based backbone, whereas for the ResNet-based backbone, we always utilize a full fine-tuning strategy.

\begin{algorithm}[!t]
\caption{Pseudo-Code of {\ours}}
\label{alg:algorithm}
\begin{flushleft}
\textbf{Inputs}: \\
Training dataset $\mathcal{D}_{\text{train}} = \{(\boldsymbol{x}^k, \boldsymbol{y}^k)\}_{k=1}^N$, CLIP encoders $E_I(\cdot)$ and $E_T(\cdot)$, GCN $\mathcal{G}(\cdot)$ \\
\textbf{Outputs}: \\
Predicted probability $\boldsymbol{p}_{*}$ for test instance $\boldsymbol{x}_{*}$ \\
\end{flushleft}
\begin{algorithmic}[1]
\STATE \textbf{Training Phase}:
\STATE Initialize soft prompts $\{\boldsymbol{t}_c|_1^M\}_{c=1}^C$ as per Eq.~(\ref{eq:init_tp})
\FOR{epoch $t = 1$ to $T$}
    \FOR{batch $\{(\boldsymbol{x}^k, \boldsymbol{y}^k)\}_{k=1}^B$ in $\mathcal{D}_{\text{train}}$}
        \STATE Derive text features $\boldsymbol{F}_t = E_T(\{\boldsymbol{t}_c|_1^M\}_{c=1}^C)$
        \STATE Obtain correlated text features $\boldsymbol{F}_{ct} = \mathcal{G}(\boldsymbol{F}_t)$
        \STATE Apply residual connection: $\boldsymbol{F}_t^* = \boldsymbol{F}_t + \boldsymbol{F}_{ct}$
        \FOR{each image $\boldsymbol{x}^k$ in batch}
            \STATE Extract visual embedding $\boldsymbol{v}^k = E_I(\boldsymbol{x}^k)$
            \STATE Evaluate prediction $p(y_c^k \mid \boldsymbol{x}^k)$ using Eq.~(\ref{eq:sampro})
        \ENDFOR
        \STATE Aggregate class frequencies $n_c = \sum_{i=1}^B y_c^i$
        \STATE Calculate re-balanced weights $r$ using Eq.~(\ref{eq:re_weight})
        \STATE Estimate class-specific biases $v_c$ using Eq.~(\ref{eq:class_bias})
        \STATE Adjust modulated probability $q_c^i$
        \STATE Assess per-image loss $\ell_{cls}$ using Eq.~(\ref{eq:cls_loss})
    \ENDFOR
    \STATE Aggregate classification loss $\mathcal{L}_{cls}$ using Eq.~(\ref{eq:full_cls_loss})
    \STATE Update model parameters $\Phi$ via full fine-tuning by default; apply PEFT only when explicitly specified.
\ENDFOR \\

\STATE \textbf{Inference Phase}:
\IF{test-time ensembling is activated}
    \STATE Resize instance $\boldsymbol{x}_{*}$ to $\boldsymbol{x}_{*}'$ sized $(224 + e) \times (224 + e)$
    \STATE Extract five crops: center $\boldsymbol{x}^{ce}$, top-left $\boldsymbol{x}^{tl}$, top-right $\boldsymbol{x}^{tr}$, bottom-left $\boldsymbol{x}^{bl}$, bottom-right $\boldsymbol{x}^{br}$ (each $224 \times 224$)
    \FOR{each crop $\boldsymbol{x}^{cr} \in \{\boldsymbol{x}^{ce}, \boldsymbol{x}^{tl}, \boldsymbol{x}^{tr}, \boldsymbol{x}^{bl}, \boldsymbol{x}^{br}\}$}
        \STATE Compute visual embedding $\boldsymbol{v}^{cr} = E_I(\boldsymbol{x}^{cr})$
        \STATE Compute crop probability $\boldsymbol{p}^{cr}$ using Eq.~(\ref{eq:sampro})
    \ENDFOR
    \STATE Ensemble probability: $\boldsymbol{p}_{*} = \frac{1}{5} \sum_{cr} \boldsymbol{p}^{cr}$
\ELSE
    \STATE Compute visual embedding $\boldsymbol{v}_{*} = E_I(\boldsymbol{x}_{*})$
    \STATE Compute predicted probability $\boldsymbol{p}_{*}$ using Eq.~(\ref{eq:sampro})
\ENDIF
\end{algorithmic}
\end{algorithm}

\subsection{End-to-End Optimization}
The proposed framework {\ours} undergoes end-to-end optimization with a classification loss $\mathcal{L}_{\text{cls}}$ that promotes similarity in probabilities among correlated labels. This mechanism cultivates semantic associations and mitigates long-tailed biases by propagating knowledge from head to tail classes. The learnable parameters encompass soft prompts and graph convolutional network weights.
To boost performance, we apply test-time ensembling across ViT-based and ResNet-based backbones. For the ViT-based backbone, we mitigate overfitting on tail classes through parameter-efficient fine-tuning, which restricts trainable parameters in the image encoder to facilitate efficient adaptation and maintain generalization. This approach improves efficacy in data-scarce settings and conforms to advancements in scalable vision-language model adaptation for imbalanced real-world datasets. \looseness=-1

To provide a comprehensive overview of the end-to-end pipeline, Algorithm \ref{alg:algorithm} outlines the pseudocode for both training and inference phases. During training, the model is optimized with a long-tailed multi-label classification loss, while the GCN module captures inter-class correlations and PEFT enables parameter-efficient adaptation. In inference, TTE can be optionally applied to further improve performance.

\begin{table*}[!t]
\centering
\caption {Performance (mAP\%) comparison with conventional and adapted methods using ResNet-50.}
    \begin{tabular}{cccccccccc}
        \toprule
        \multirow{2.5}{*}{Category}   &   \multirow{2.5}{*}{Algorithm} & \multicolumn{4}{c}{VOC-LT}     & \multicolumn{4}{c}{COCO-LT}    \\
        \cmidrule(lr){3-6}\cmidrule(l){7-10}
         &     & Total & Head  & Medium & Tail  & Total & Head  & Medium & Tail  \\
        \midrule
        \multirow{3}{*}{MLC} 
          &  ML-GCN~\cite{chen2019multi} & 68.92 & 70.14 & 76.41 & 62.39 & 44.24 & 44.04 & 48.36 & 38.96 \\
          &  Focal Loss~\cite{lin2020focal} & 73.88 & 69.41 & 81.43  & 71.56 & 49.46 & 49.80 & 54.77  & 42.14 \\
          &  ASL~\cite{RidnikBZNFPZ21} & 78.31 & 71.12 & 84.95 & 78.71 & 54.35 & 50.59 & 58.76 & 51.82 \\ \hline
        \multirow{7}{*}{LT-SLC} 
          &  ERM & 70.86 & 68.91 & 80.20  & 65.31 & 41.27 & 48.48 & 49.06  & 24.25 \\
          &  RS~\cite{shen2016relay}  & 75.38 & 70.95 & 82.94  & 73.05 & 46.97 & 47.58 & 50.55  & 41.70 \\
          &  RW  & 74.70 & 67.58 & 82.81  & 73.96 & 42.27 & 48.62 & 45.80  & 32.02 \\
          &  OLTR~\cite{liu2019large} & 71.02 & 70.31 & 79.80 & 64.95 & 45.83 & 47.45 & 50.63 & 38.05 \\ 
          &  LDAM~\cite{cao2019learning} & 70.73 & 68.73 & 80.38 & 69.09 & 40.53 & 48.77 & 48.38 & 22.92 \\
          &  CB Focal~\cite{cui2019class}   & 75.24 & 70.30  & 83.53  & 72.74 & 49.06 & 47.91 & 53.01  & 44.85 \\
          &  BBN~\cite{zhou2020bbn} & 73.37 & 71.31 & 81.76 & 68.62 & 50.00 & 49.79 & 53.99 & 44.91 \\ 
        \hline
        \multirow{10}{*}{LT-MLC} 
          &  DB ~\cite{wu2020distribution} &  78.65 & 73.16 & 84.11 & 78.66 & 52.53 & 50.25 & 56.33 & 49.54 \\ 
          &  DB Focal~\cite{wu2020distribution}  & 78.94 & 73.22 & 84.18  & 79.30  & 53.55 & 51.13 & 57.05  & 51.06 \\
          &  URS~\cite{guo2021long}  & 81.44 & 75.68 & 85.53  & 82.69 & 56.90 & 54.13 & 60.59  & 54.47 \\
          &  CAEN~\cite{chen2023class} & 81.61 &74.00 &85.35 &85.28 &57.64 &52.37 &61.18 &57.63\\
          &  MFM~\cite{ZhangLZOTZ23}  & 79.64  & 66.32  & 84.69  & 85.83  & 55.25  & 48.71  & 58.24  & 57.08 \\
          &  DR Loss~\cite{LinPCXQC24}  & 78.01  & 72.19  & 83.83  & 77.69  & 54.10  & 50.27  & 57.00  & 53.76 \\
          &  MLC-NC~\cite{TaoLWZCLHC25}  & 84.37 & 72.75 & 88.15 & 90.31 & 60.52 & 49.69 & 64.94 & 64.21 \\
          &  \gray  {\ours} & \gray 87.46  &  \gray 81.13  &  \gray 88.73 &  \gray 91.78      & \gray 64.44 & \gray 57.70  & \gray 67.72 & \gray 66.05 \\
         & \gray {\ours} + TTE & \gray \bf 88.42 & \gray \bf82.45 & \gray \bf89.10 & \gray \bf92.38 & \gray \bf 65.75 & \gray \bf58.54 & \gray \bf 68.09 & \gray \bf69.00\\
        \bottomrule 
    \end{tabular}
\label{tab:map_main}
\end{table*}

\section{Experiments}
\label{sec:experiments}
\subsection{Experiment Settings}
\noindent
\textbf{Datasets.}
To evaluate the effectiveness of our proposed approach for long-tailed multi-label visual recognition, we conduct experiments on two widely adopted benchmark datasets: VOC-LT and COCO-LT~\cite{wu2020distribution}. These datasets are artificially constructed by subsampling from established multi-label recognition benchmarks, Pascal VOC~\cite{everingham2015pascal} and MS COCO~\cite{lin2014microsoft}, respectively, to simulate the long-tailed distribution prevalent in real-world scenarios. Specifically, all classes are partitioned into three groups based on the number of training samples per class: head, medium, and tail.

The VOC-LT dataset is derived from the 2012 train-val split of Pascal VOC, following a Pareto distribution for sampling imbalance. It comprises 1142 training images across 20 object classes, with the number of samples per class varying from 4 to 775. The class distribution ratio is 6:6:8 for head, medium, and tail classes, where head classes have more than 100 samples, medium classes have between 20 and 100 samples, and tail classes have fewer than 20 samples. For evaluation, we use the full VOC 2007 test set, which contains 4952 images.

Similarly, the COCO-LT dataset is sampled from the MS COCO 2017 training set, resulting in 1909 training images and 80 object classes, with samples per class ranging from 6 to 1128. The class distribution follows a 22:33:25 ratio for head, medium, and tail classes, adhering to the same criteria as VOC-LT. The test set consists of the entire 5000 images from the MS COCO 2017 validation set. 

\noindent
\textbf{Evaluation Metrics.}
Consistent with prior studies on long-tailed multi-label visual recognition \cite{wu2020distribution, guo2021long, xia2023lmpt}, we adopt the mean average precision (mAP) as the primary evaluation metric. This metric computes the average precision across all classes and images, providing a comprehensive assessment of model performance under imbalanced conditions. To gain deeper insights into the handling of tail classes, we also report mAP scores stratified by head, medium, and tail groups.

\noindent
\textbf{Comparative Methods.}
To assess the efficacy of our proposed {\ours}, we benchmark it against a diverse set of state-of-the-art approaches categorized into three groups: (1) general multi-label classification (MLC) methods, which serve as foundational baselines; (2) long-tailed single-label classification (LT-SLC) methods, adapted to the multi-label setting to handle imbalance; and (3) specialized long-tailed multi-label classification (LT-MLC) methods, directly tailored for the challenges of long-tailed distributions in multi-label scenarios. Additionally, we include comparisons with CLIP-based methods to highlight the advantages of leveraging VLMs. \looseness=-1

For MLC methods, we include ML-GCN~\cite{chen2019multi}, which exploits label correlations via graph convolutional networks; Focal loss~\cite{lin2020focal}, designed to focus on hard examples; and ASL~\cite{RidnikBZNFPZ21}, an asymmetric loss for improved multi-label performance.
For LT-SLC methods, we adapt several prominent techniques: empirical risk minimization (ERM), re-weighting (RW), and re-sampling (RS)~\cite{shen2016relay}; OLTR~\cite{liu2019large}, which uses dynamic meta-embedding; LDAM~\cite{cao2019learning}, a label-distribution-aware margin loss; class-balanced (CB) Focal loss~\cite{cui2019class}, combining class-balancing with focal loss; and BBN~\cite{zhou2020bbn}, a cumulative learning strategy.
For LT-MLC methods, we compare with distribution-balanced (DB) loss and DB Focal loss~\cite{wu2020distribution}, which address distribution imbalance; URS~\cite{guo2021long}, utilizing uniform and re-balanced sampling collaboratively; CAEN~\cite{chen2023class}, a class-aware enhancement network; MFM~\cite{ZhangLZOTZ23}, a multi-feature modulation approach; DR Loss~\cite{LinPCXQC24}, a distributionally robust loss; and the recent MLC-NC~\cite{TaoLWZCLHC25}, which incorporates noise-contrastive estimation for multi-label learning. 
Furthermore, we evaluate against LMPT~\cite{xia2023lmpt}, a CLIP-based prompt tuning method for long-tailed multi-label recognition.

\noindent
\textbf{Implementation Details.}
Our framework is built upon vision-language models, specifically initializing the visual-text encoders with pre-trained CLIP weights. We employ two variants for ablation and comparison: ResNet-50-based CLIP~\cite{he2016deep} and ViT-Base/16-based CLIP \cite{dosovitskiy2021image}. For fairness, all CLIP-based methods initialize the prompt with the default hand-crafted template ``a photo of a [CLS]''. During fine-tuning, the text encoder parameters are frozen to preserve the rich semantic representations, while the image encoder and learnable prompts are updated. For the parameter-efficient fine-tuning strategy, only a subset of the image encoder parameters is tuned via adapters to minimize computational overhead.

All experiments utilize the Adam optimizer with a batch size of 32. Learning rates are empirically set to $5 \times 10^{-4}$ for COCO-LT and $4 \times 10^{-5}$ for VOC-LT. For the graph convolutional network component, learning rates of $1 \times 10^{-3}$ and $1 \times 10^{-4}$ are used for COCO-LT and VOC-LT, respectively. Hyperparameters in the classification loss align with those in~\cite{wu2020distribution}, with the exception of $\beta = 0.01$ to better balance the contributions from head and tail classes. Implementations are carried out in PyTorch 2.0.1 on an NVIDIA GeForce RTX 3090 GPU\footnote{The code will be publicly available.}. \looseness=-1

\begin{table*}[!t]
\centering
\caption {Performance (mAP\%) comparison with CLIP-based methods using ResNet-50 or ViT-Base/16.}
    \begin{tabular}{ccccccccc}
        \toprule
        \multirow{2.5}{*}{Algorithm}    & \multicolumn{4}{c}{VOC-LT}     & \multicolumn{4}{c}{COCO-LT}    \\
        \cmidrule(lr){2-5} \cmidrule(l){6-9}
            & Total & Head  & Medium & Tail  & Total & Head  & Medium & Tail  \\
        \midrule
        Backbone: \textit{\textbf{ResNet-50}} &&&&&&&& \\
        Zero-Shot CLIP~\cite{radford2021learning}  & 84.30  & 63.60  & 88.03  & 97.03  & 56.19  & 35.73  & 60.52  & 68.45 \\
        CoOp~\cite{zhou2022learning}  & 81.34  & 65.10  & 81.54  & 93.37  & 54.94  & 38.06  & 56.67  & 67.51 \\
        CoCoOp~\cite{zhou2022conditional}  & 78.63  & 64.33  & 80.51  & 87.94  & 46.02  & 36.02  & 50.57  & 48.82 \\
        LMPT~\cite{xia2023lmpt} & 85.44 & 66.62 & 88.11 & \bf97.86 & 58.97 & 41.87 & 61.60 & \bf69.60 \\
        \gray  {\ours} & \gray 87.46  & \gray 81.13  &  \gray 88.73 &  \gray 91.78      & \gray 64.44 & \gray 57.70  & \gray 67.72 & \gray 66.05 \\
        \gray {\ours} + TTE & \gray \bf 88.42 & \gray \bf82.45 & \gray \bf89.10 & \gray 92.38 & \gray \bf 65.75 & \gray \bf58.54 & \gray \bf 68.09 & \gray 69.00\\
        \midrule 
        Backbone: \textit{\textbf{ViT-Base/16}} &&&&&&&& \\
        Zero-Shot CLIP~\cite{radford2021learning}  & 85.77 & 66.52 & 88.93  & 97.83 & 60.17 & 38.52 & 65.06  & 72.28 \\
        CoOp~\cite{zhou2022learning} & 86.02 & 67.71 & 88.79 & 97.67 & 60.68 & 41.97 & 63.18 &  73.85 \\
        CoCoOp~\cite{zhou2022conditional} & 84.47 & 64.58 & 87.82 & 96.88 & 61.49 & 39.81 & 64.63 & 76.42 \\
        LMPT~\cite{xia2023lmpt}   & 87.88 & 72.10 & 89.26 & \bf98.49 & 66.19 & 44.89  & 69.80  & \bf79.08 \\
        \gray {\ours} & \gray 89.98 & \gray 83.69 & \gray 90.30 & \gray 94.53 & \gray 72.18 & \gray 64.11 & \gray 76.75 & \gray 73.86\\
        \gray {\ours} + TTE & \gray 90.43 & \gray 84.71 & \gray 90.73 & \gray 94.48 & \gray 72.69 & \gray 64.82 & \gray 76.78 & \gray 73.88 \\
        \gray {\ours} + PEFT & \gray 92.56 & \gray 85.40 & \gray 93.30 & \gray 97.36 &  \gray 75.34 & \gray 66.63 & \gray 78.44 & \gray 78.74 \\
        \gray {\ours} + TTE + PEFT & \gray \bf 93.03 & \gray \bf85.86 & \gray \bf94.05 & \gray 97.64 & \gray \bf 76.36 & \gray \bf68.87 & \gray \bf 79.46 & \gray 78.87\\
        \bottomrule
    \end{tabular}
\label{tab:map_clip}
\end{table*}

\subsection{Results on the VOC-LT and COCO-LT Datasets}
To rigorously evaluate the proposed {\ours}, we present comprehensive performance comparisons on VOC-LT and COCO-LT datasets. Table~\ref{tab:map_main} summarizes the mean average precision (mAP) results across various categories of methods using a ResNet-50 backbone, while Table~\ref{tab:map_clip} extends this analysis to CLIP-based approaches with ResNet-50 and ViT-Base/16 backbones. These results highlight the superior capability of {\ours} in addressing the challenges of long-tailed multi-label visual recognition, achieving new state-of-the-art performance on both datasets. \looseness=-1

\noindent
\textbf{Comparative Performance Analysis with Conventional and Adapted Methods Using ResNet-50.}
As shown in Table~\ref{tab:map_main}, {\ours} attains an overall mAP of 87.46\% on VOC-LT and 64.44\% on COCO-LT, surpassing the previous best LT-MLC method, MLC-NC \cite{TaoLWZCLHC25}, by 3.09\% and 3.92\%, respectively. This improvement is particularly pronounced across class groups: on VOC-LT, {\ours} boosts mAP by 8.38\% for head classes, 0.58\% for medium classes, and 1.47\% for tail classes compared to MLC-NC; on COCO-LT, the gains are 8.01\% for head classes, 2.78\% for medium classes, and 1.84\% for tail classes. Such balanced enhancements underscore the method's effectiveness in mitigating the dominance of head classes while elevating the recognition of underrepresented tail classes, a core issue in long-tailed distributions.
Incorporating the test-time ensembling (TTE) technique further boosts performance, achieving mAP scores of 88.42\% on VOC-LT and 65.75\% on COCO-LT, corresponding to gains of 0.96\% and 1.31\% over the base {\ours}. TTE refines the semantic alignment between visual features and text prompts, enabling more precise multi-label predictions, especially in scenarios with sparse training data for tail classes. \looseness=-1

Compared to general MLC methods like ML-GCN~\cite{chen2019multi}, Focal Loss~\cite{lin2020focal}, and ASL~\cite{RidnikBZNFPZ21}, which achieve mAPs ranging from 68.92\% to 78.31\% on VOC-LT and 44.24\% to 54.35\% on COCO-LT, {\ours} demonstrates substantial gains (up to 19.5\% on VOC-LT and 21.51\% on COCO-LT). While these baselines are effective on balanced or single-labeled datasets, their performance deteriorates under long-tailed multi-label visual recognition owing to their inability to explicitly mitigate class skewness.
LT-SLC methods, e.g., ERM, RS~\cite{shen2016relay}, RW, OLTR~\cite{liu2019large}, LDAM~\cite{cao2019learning}, CB Focal~\cite{cui2019class}, BBN~\cite{zhou2020bbn}, yield mAPs of 70.73\%--75.38\% on VOC-LT and 40.53\%--50.00\% on COCO-LT. These baselines, while effective for balanced datasets or single-label tasks, underperform in long-tailed multi-label visual recognition due to their limited capacity to explicitly counteract class skewness and preserve inter-label correlations under severe data imbalance.
Compared to LT-MLC methods, e.g., DB~\cite{wu2020distribution}, DB Focal~\cite{wu2020distribution}, URS~\cite{guo2021long}, CAEN~\cite{chen2023class}, MFM~\cite{ZhangLZOTZ23}, DR Loss~\cite{LinPCXQC24}, MLC-NC~\cite{TaoLWZCLHC25}, {\ours} consistently outperforms, with gains of 3.05\%--9.45\% on VOC-LT and 4.23\%--23.91\% on COCO-LT in total mAP. This superiority stems from leveraging pre-trained vision-language models and graph convolutional networks, which provide robust, semantically rich representations that bridge the gap between head and tail classes. \looseness=-1

\noindent
\textbf{Comparative Performance Analysis with CLIP-Based Methods.}
Shifting to CLIP-based methods in Table~\ref{tab:map_clip}, zero-shot CLIP \cite{radford2021learning} establishes a strong baseline of 84.30\% on VOC-LT and 56.19\% on COCO-LT with ResNet-50, benefiting from large-scale pre-training. Prompt-tuning methods like CoOp~\cite{zhou2022learning} and CoCoOp~\cite{zhou2022conditional} refine this baseline. However, LMPT~\cite{xia2023lmpt} achieves 85.44\% on VOC-LT and 58.97\% on COCO-LT. In contrast, {\ours} fine-tunes the image encoder and employs sigmoid-based probability estimation to align with multi-label paradigms, unlike LMPT's softmax. Consequently, {\ours} surpasses LMPT by 2.02\% and 5.47\% in total mAP. While LMPT excels marginally on tail classes in some cases, as evidenced by 97.86\% versus 91.78\% on VOC-LT tail classes, {\ours} provides more balanced improvements, particularly on head and medium classes, as evidenced by 81.13\% versus 66.62\% on head classes.

Using ViT-Base/16, {\ours} reaches 89.98\% on VOC-LT and 72.18\% on COCO-LT, approximating 96.8\% and 95.0\% of full fine-tuning performance while being more parameter-efficient. Incorporating TTE adds ~0.5\% gains, and combining with parameter-efficient fine-tuning (PEFT) via AdaptFormer~\cite{chen2022adaptformer} yields remarkable results: 93.03\% on VOC-LT and 76.36\% on COCO-LT with TTE and PEFT, outperforming LMPT by 5.15\% and 10.17\% overall. This configuration enhances tail class mAP by up to 4.83\% over base {\ours}, demonstrating PEFT's role in scalable adaptation without overfitting.
These results collectively validate {\ours} as a robust framework that unleashes vision-language models' potential for long-tailed multi-label recognition, offering balanced, high-performance predictions across class distributions and setting new benchmarks for future research.

\subsection{Ablation Studies}
To dissect the contributions of individual components in our proposed framework and validate design choices, we conduct comprehensive ablation studies on the VOC-LT and COCO-LT datasets using the ViT-Base/16 backbone (unless otherwise specified). These experiments elucidate the impact of each module on addressing long-tailed multi-label visual recognition challenges, providing insights into how vision-language models can be effectively harnessed for improved performance under data imbalance.

\begin{table*}[!t]
\centering
\caption {Ablation analysis(mAP\%) on different components using ViT-Based/16.}
\setlength{\tabcolsep}{1.5mm}{
    \begin{tabular}{ccccccccccccc}
        \toprule 
        & \multicolumn{4}{c}{Components}   & \multicolumn{4}{c}{VOC-LT} & \multicolumn{4}{c}{COCO-LT} \\ 
        \cmidrule(lr){2-5}  \cmidrule(lr){6-9}   \cmidrule(l){10-13}
        & $\mathcal{L}_{cls}$  & GCN& TTE & PEFT  & Total & Head & Medium & Tail  & Total & Head  & Medium & Tail \\
        \midrule 
        Baseline  &  & & &  & 88.21 & 84.21 & 87.42 & 91.80 & 69.65 & 64.77 & 71.20  & 69.28\\
        &  \Checkmark&  && & 88.94 &81.86 & 90.07 & 93.41 & 71.12 & 64.18 & 74.51 & 72.74\\
         {\ours} &  \Checkmark & \Checkmark &  & & 89.98 & 83.69 & 90.30 & 94.53 & 72.18 & 64.11 & 76.75 & 73.86\\
        &  \Checkmark & \Checkmark & \Checkmark & & 90.43 & 84.71 & 90.73 & 94.48 & 72.69 & 64.82 & 76.78 & 73.88 \\
        &  \Checkmark & \Checkmark & & \Checkmark    &92.56 & 85.40 & 93.30 & 97.36 &  75.34 & 66.63 & 78.44 & 78.74\\
        &  \Checkmark & \Checkmark & \Checkmark & \Checkmark&\bf 93.03 & \bf85.86 & \bf94.05 & \bf97.64 & \bf 76.36 & \bf68.87 &\bf 79.46 & \bf78.87\\
        \bottomrule
    \end{tabular}
} 
\label{tab:components}
\end{table*}

\subsubsection{\textbf{Components Analysis}}
We systematically ablate key components of our framework, starting from a vanilla baseline and progressively incorporating modules. As reported in Table~\ref{tab:components}, the incremental improvements highlight the critical and complementary role of each component.

The vanilla baseline employs full fine-tuning of the CLIP image encoder with Focal loss, yielding mAP of 88.21\% on VOC-LT and 69.65\% on COCO-LT. Notably, this surpasses the prior state-of-the-art LMPT~\cite{xia2023lmpt} that achieved 87.88\% on VOC-LT and 66.19\% on COCO-LT, highlighting the effectiveness of full fine-tuning VLMs. However, a notable degradation occurs in tail class performance that drops by approximately 7.0\% relative to zero-shot CLIP. This drop is attributable to the long-tailed distribution and multi-label setting, which disrupts pre-trained correspondences between images and labels.

Incorporating the long-tailed multi-label classification loss $\mathcal{L}_{cls}$ enhances overall mAP to 88.94\% on VOC-LT and 71.12\% on COCO-LT, with substantial tail class gains of 1.61\% and 3.46\%, respectively. This loss explicitly counters class imbalance by re-weighting contributions, promoting equitable learning across head, medium, and tail classes.
The GCN semantic module further refines text features by modeling inter-label relationships, yielding 1.04\% and 1.06\% mAP improvements overall on VOC-LT and COCO-LT, respectively. By propagating semantic information through a graph constructed from prompt similarities, GCN enriches representations and particularly benefits correlated labels in sparse tail regions.

Adding the TTE module on top of the classification loss and GCN provides consistent refinements, improving mAP by 0.45\% on VOC-LT and 0.51\% on COCO-LT. TTE optimizes text prompts to enhance visual-text alignment, addressing subtle mismatches in long-tailed scenarios where tail classes often lack diverse representations. This module contributes modestly but crucially to maintaining semantic coherence, as evidenced by balanced gains across head, medium, and tail classes, with tail improvements of approximately 0.05\% on VOC-LT and 0.02\% on COCO-LT relative to the prior configuration.
Integrating PEFT via AdaptFormer, when added separately to the classification loss and GCN, yields substantial boosts: 2.58\% on VOC-LT and 3.16\% on COCO-LT in overall mAP. PEFT's low-rank adaptations enable efficient parameter updates, preserving generalization while adapting to data imbalance. This is particularly evident in tail class gains of 2.83\% on VOC-LT and 4.88\% on COCO-LT, demonstrating its efficacy in mitigating overfitting on sparse samples.
Finally, combining all components, including TTE and PEFT, achieves the highest performance with mAP scores of 93.03\% on VOC-LT and 76.36\% on COCO-LT. The synergy between TTE and PEFT is notable, as their joint application adds an additional 0.47\% on VOC-LT and 1.02\% on COCO-LT beyond the PEFT-only variant, underscoring how prompt optimization complements parameter-efficient fine-tuning. 

Collectively, these ablation studies confirm that each component addresses distinct aspects of the long-tailed multi-label visual recognition problem, such as imbalance mitigation, semantic modeling, alignment refinement, and efficient adaptation, culminating in state-of-the-art results that significantly outperform baselines across all class groups.

\begin{figure}[!t]
    \centering
    \begin{overpic}[width=7cm]{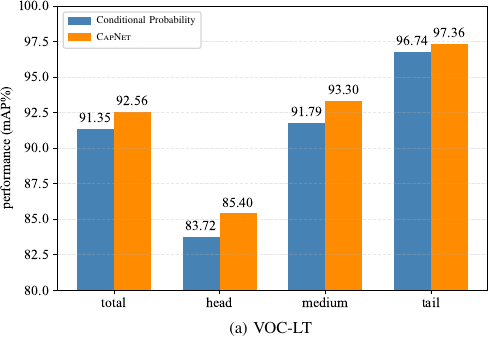}
    \end{overpic}
    \vspace{-0.1cm}
    \begin{overpic}[width=7cm]{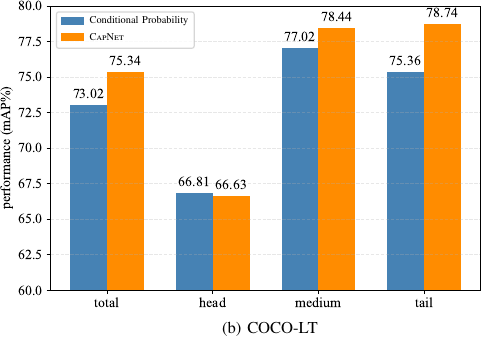}
    \end{overpic}
    \caption{Comparison of different correlation matrices on VOC-LT and COCO-LT datasets.}
    \label{fig:dcm}
\end{figure}

\subsubsection{\textbf{Correlation Matrix Analysis}}
In this part, we conduct experiments using ViT-Base/16 as the backbone, AdaptFormer as the PEFT module, and without the TTE technique.

\noindent
\textbf{Effects of Different Correlation Matrices.}
In this section, we employ the ViT-Base/16 with AdaptFormer as the PEFT module, and refrain from using TTE techniques in experiments to observe the impact of different correlation matrices on benchmark datasets. Here, we validate two strategies: (1) modeling the correlation matrix as conditional probabilities, as adopted in ML-GCN~\cite{chen2019multi}; (2) our method, which constructs the matrix based on the cosine similarity between the prompt templates of each class. As illustrated in Fig.~\ref{fig:dcm}, our approach achieves significant performance improvements, demonstrating its effectiveness in capturing label correlation information.

On VOC-LT, our method improves mAP by 1.21\% overall, with gains of 1.68\% on head classes, 1.51\% on medium classes, and 0.62\% on tail classes. These improvements indicate that cosine similarity captures semantic relationships more robustly and is less affected by data imbalance. On COCO-LT, our approach yields a 2.32\% overall increase. Notably, medium and tail classes benefit most, with 1.42\% and 3.38\% gains, respectively. The performance on the head classes in COCO-LT shows a minor decrease of 0.18\% mAP compared to the conditional probability approach. We conjecture that this arises because conditional probabilities heavily rely on empirical co-occurrence statistics derived from the training data, which are abundant and reliable for head classes due to their high frequency. As a result, that method can overfit to these dominant patterns, yielding marginally better results on frequent labels.
This analysis reveals that prompt-based similarities leverage pre-trained linguistic knowledge, providing a more balanced representation across class frequencies. In long-tailed settings, where tail classes have sparse co-occurrences, semantic embeddings mitigate bias, leading to enhanced tail performance. Furthermore, the graph constructed this way integrates seamlessly with GCNs, propagating information effectively from head to tail classes. \looseness=-1

\begin{figure}[!t]
    \centering
    \begin{overpic}[width=7cm, height=4.9cm]{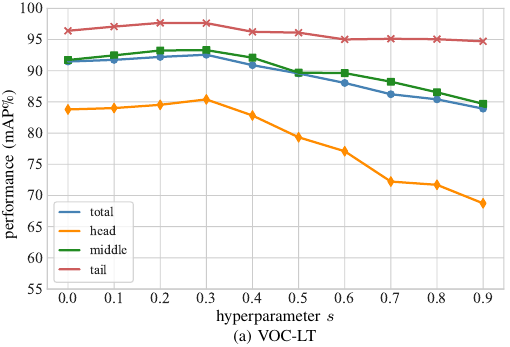}
    \end{overpic}
    \vspace{-0.2cm}
    \begin{overpic}[width=7cm, height=4.9cm]{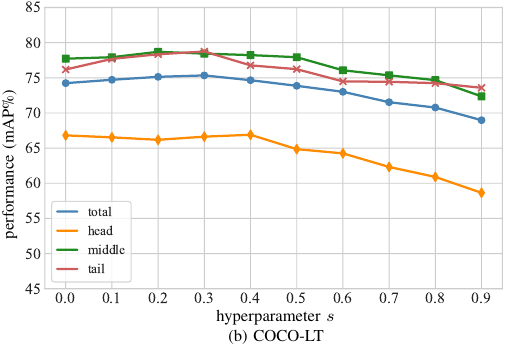}
    \end{overpic}
    \caption{Performance (mAP\%) comparisons with different hyperparameter $s$ on COCO-LT and VOC-LT datasets.}
    \label{fig:ablation_s}
\end{figure}

\noindent
\textbf{Effects of Different $s$ for Correlation Matrix.} 
To investigate the impact of varying the hyperparameter $s$ in Eq.~(\ref{eq:co_m1}) on mean average precision, we evaluate $s$ across the set $\{0, 0.1, 0.2, \dots, 0.9\}$, as illustrated in Fig.~\ref{fig:ablation_s}. This analysis highlights the critical role of balancing the contribution from a node's own features and those of its neighboring nodes during feature updates in the graph convolutional network.

On the VOC-LT dataset,  the total mean average precision commences at 91\% for $s=0$ and diminishes to 85\% at $s=0.9$. The head classes display a slight rise from 84\% to 86\% at $s=0.1$ before falling to 69\%, indicating minor gains from limited neighborhood integration but vulnerability to over-smoothing. The middle classes maintain high performance around 92\% initially, then drop to 85\%. The tail classes achieve the most pronounced benefits, rising from 96\% at $s=0$ to nearly 97\% at $s=0.2$ and sustaining elevated levels until $s=0.3$, after which they decrease to 95\%. This demonstrates that tail classes particularly capitalize on correlation-based feature propagation at optimal $s$ values.
Similar trends emerge on the COCO-LT dataset, where the total mean average precision starts at approximately 75\% when $s=0$ and gradually declines to around 70\% as $s$ approaches 0.9, indicating a general sensitivity to higher values of $s$. The head classes, which are frequent and data-rich, exhibit a stable performance around 65\% for small $s$ before a moderate decline to about 60\%, suggesting that excessive incorporation of neighborhood information introduces noise without substantial benefits. In contrast, the middle classes begin at 78\% and decrease steadily to 73\%, reflecting a consistent but controlled degradation. Notably, the tail classes, which are rare and benefit from label correlations, show an initial improvement from 76\% at $s=0$ to a peak near 78\% around $s=0.3$, followed by a decline to 74\% at higher $s$. This pattern underscores how moderate $s$ values enable tail classes to leverage information from correlated head classes, enhancing their recognition, while larger $s$ leads to over-smoothing and performance loss. \looseness=-1

Across both datasets, $s=0.3$ yields the best overall performance, as it optimally balances the infusion of correlated label information to bolster tail class accuracy without excessively diluting node-specific features, which would otherwise cause over-smoothing. When $s$ is too small, labels receive insufficient contextual support from correlated counterparts, limiting improvements in underrepresented classes. Conversely, excessively large $s$ homogenizes features across the graph, degrading discrimination across all class partitions.

\begin{table*}[!t]
\caption{Comparison (parameter and mAP\%) of our method with different parameter-efficient fine-tuning methods.}\label{tabv}
\centering
    \begin{tabular}{l| c | c c c c | c c c c}
        \toprule
        \multirow{2}{*}{\bf Methods} & \multirow{2}{*}{\bf Param.} & \multicolumn{4}{c|}{\bf VOC-LT} & \multicolumn{4}{c}{\bf COCO-LT} \\ 
        & &\bf Total &\bf Head &\bf Medium &\bf Tail  &\bf Total &\bf Head &\bf Medium &\bf Tail \\
        \midrule
        
        Full fine-tuning &\bf 91.34M  & 89.98 & 83.69 & 90.30 & 94.53 & 72.18 & 64.11 & 76.75 & 73.86 \\
        VPT-shallow~\cite{jia2022visual}&\bf 6.30M  & 79.50 & 69.53 & 80.94  & 85.89 & 63.34 & 53.34 & 66.71 & 67.72 \\
        VPT-deep~\cite{jia2022visual}&\bf 6.39M  & 90.52 & 82.64 & 90.98 & 96.08  & 74.64 & 66.40 & 77.93 & 77.55 \\
        \midrule
        AdaptFormer-2~\cite{chen2022adaptformer}&\bf 6.35M  & 91.22 & 83.44 & 91.85 & 96.59  & 74.24 & 65.17 & 78.22 & 76.97 \\
        AdaptFormer-4~\cite{chen2022adaptformer}&\bf 6.39M  & 91.37 & 84.05 & 92.31 & 96.14  & 74.74 & 66.16 & 78.27 & 77.67 \\
        AdaptFormer-16~\cite{chen2022adaptformer}&\bf 6.61M  &\bf 92.56 &\bf 85.40 & 93.30 & 97.36  & 75.27 &\bf 67.30 &78.49 & 78.04 \\
        AdaptFormer-32~\cite{chen2022adaptformer}&\bf 6.91M  & 92.47 & 85.18 & 93.16 &\bf 97.43  &\bf 75.34 & 66.63 & 78.44 &\bf 78.74 \\
        AdaptFormer-64~\cite{chen2022adaptformer}&\bf 7.49M  & 92.45 & 85.39 &\bf 93.36 & 97.08  & 75.23 & 66.72 & \bf78.59 & 78.27 \\
        \bottomrule
    \end{tabular}
\label{tab:peft}
\end{table*}

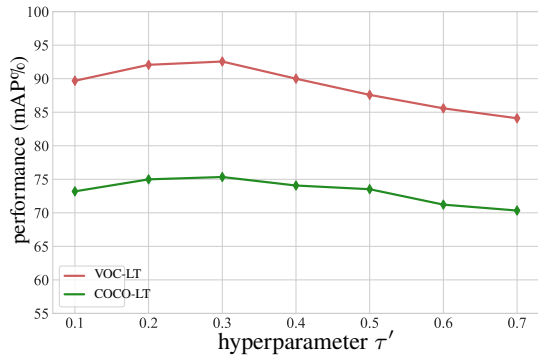
\begin{figure}[!t]
\centering
    \begin{overpic}[width=7cm]{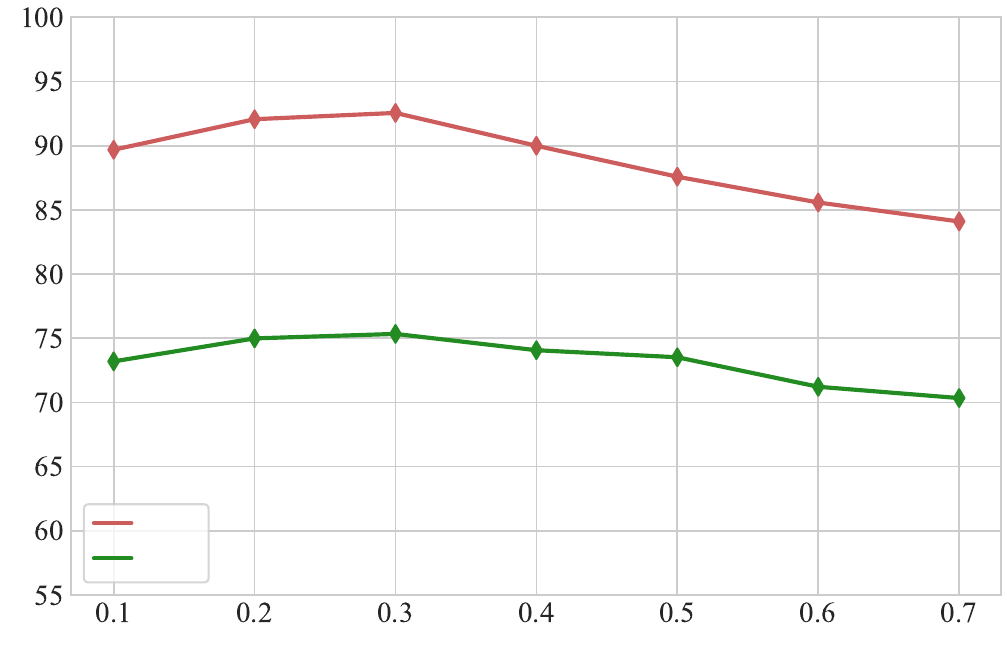}
  \put(-10, 190) {\begin{turn}{90} \footnotesize performance (mAP\%)\end{turn}}
  \put(385, -10) {\small hyperparameter $\tau^{\prime}$}
  \put(150, 125) {\tiny VOC-LT}
  \put(150, 87) {\tiny COCO-LT}
    \end{overpic}
\caption{Performance (total mAP\%) comparisons comparisons with different values of $\tau^{\prime}$ on COCO-LT and VOC-LT datasets.}
\label{fig:tau}
\end{figure}

\noindent
\textbf{Effects of Different Hyper-parameter $\tau^{\prime}$.} 
To examine the influence of the threshold $\tau^{\prime}$ in Eq.~(\ref{eq:co_m2}) on the smoothness of the distribution within graph $\mathcal{G}$, we evaluate performance across a range of $\tau^{\prime}$ values from 0.1 to 0.7, as presented in Fig.~\ref{fig:tau}. This hyperparameter governs the degree of smoothing applied to label correlations, and higher values lead to denser connections that may cause over-smoothing.

On the VOC-LT dataset, total mAP initiates at approximately 90\% for $\tau^{\prime}=0.1$, rises to a peak of about 93\% at $\tau^{\prime}=0.3$, and then declines steadily to around 85\% at $\tau^{\prime}=0.7$. This trend indicates that moderate $\tau^{\prime}$ enhances label correlations effectively.
For the COCO-LT dataset, performance begins at 73\% for $\tau^{\prime}=0.1$, reaches an optimum of 75\% at $\tau^{\prime}=0.3$, and decreases to 71\% at $\tau^{\prime}=0.7$. The initial improvement suggests that increasing graph connectivity aids in addressing class imbalance.
Although the figure presents aggregate performance, the optimal $\tau^{\prime}=0.3$ on both datasets strikes an effective balance.

\subsubsection{\textbf{Analysis of Parameter-efficient Fine-tuning Methods}}
To identify the optimal fine-tuning strategy for our method, we evaluate full fine-tuning alongside parameter-efficient approaches, including VPT-shallow, VPT-deep, and AdaptFormer, on both VOC-LT and COCO-LT datasets. For VPT variants, we set the prompt length to 10 based on prior performance assessments. Results in Table~\ref{tab:peft} demonstrate that AdaptFormer consistently outperforms other methods across all metrics, achieving superior mAP with fewer tunable parameters. Notably, 6.29M parameters originate from the graph convolutional network module, highlighting the efficiency gains from targeted adaptation. \looseness=-1

Among the methods, VPT-shallow yields the lowest performance, with 79.50\% on VOC-LT and 63.34\% on COCO-LT, indicating insufficient depth in adaptation for long-tailed distributions. VPT-deep improves substantially to 90.52\% and 74.64\%, respectively, yet falls short of AdaptFormer variants. Full fine-tuning attains 89.98\% on VOC-LT but requires extensive parameters, underscoring the trade-off between capacity and efficiency. Focusing on class partitions, AdaptFormer enhances tail class recognition markedly; for instance, AdaptFormer-32 reaches 97.43\% on VOC-LT tail classes and 78.74\% on COCO-LT, surpassing full fine-tuning by 2.90\% and 4.88\%, respectively. Head and medium classes also benefit, with gains reflecting better exploitation of label correlations in imbalanced settings. We further examine the impact of the middle dimension in AdaptFormer, which modulates the number of introduced parameters, across $\{2, 4, 16, 32, 64\}$. Performance improves with increasing dimension up to 16 on VOC-LT (92.56\% total) and 32 on COCO-LT (75.34\% total), after which it plateaus or slightly declines due to potential overfitting or diminished returns. This suggests an optimal balance where sufficient capacity captures semantic alignments without excessive complexity, particularly aiding tail classes in multi-label visual recognition. \looseness=-1

\begin{table}[!t]
    \centering
    \caption{Effects of the prompt learning on VOC-LT.}
    \begin{tabular}{ccc}
        \toprule
         Length & Initialization &  mAP\% \\
         \midrule
          4 & ``a photo of a [CLS]'' & \bf92.56  \\
          4 & Random & 91.32  \\
          8 & Random & 90.59  \\
          16 & Random & 89.12 \\
          32 & Random & 90.07\\
         \bottomrule
    \end{tabular}
    \label{tab:promptinit}
\end{table}

\subsubsection{\textbf{Analysis of the Prompt Learning}}
To examine the influence of prompt context initialization and length on model performance, we conduct experiments on the VOC-LT dataset without employing the test-time ensembling strategy. Specifically, we compare random initialization against a dataset-specific template, ``a photo of a [CLS]'', and vary the prompt context length $M$ across the set $\{4, 8, 16, 32\}$.

Results presented in Table~\ref{tab:promptinit} reveal that initialization plays a pivotal role. Employing the tailored initialization ``a photo of a [CLS]'' elevates performance to 92.56\% at $M=4$, surpassing random initialization by 1.24 percentage points. This improvement indicates that domain-aligned prompts provide a stronger starting point for optimization, enabling the model to better capture label correlations and mitigate imbalances inherent in long-tailed data. In contrast, random initialization struggles to converge effectively without such guidance, underscoring the value of semantically meaningful priors in prompt engineering for vision-language models.
Furthermore, shorter prompt lengths yield superior mean average precision, with $M=4$ achieving the highest values under both initialization schemes. Specifically, for random initialization, performance peaks at 91.32\% for $M=4$ and declines to 89.12\% for $M=16$, before a slight recovery to 90.07\% at $M=32$. This pattern suggests that concise prompts reduce noise and focus the vision-language model on essential semantic alignments, particularly in multi-label scenarios with long-tailed distributions. Longer prompts may introduce extraneous tokens that dilute the embedding space, leading to suboptimal feature representations.
These findings emphasize the importance of compact, informed prompts in enhancing recognition accuracy, offering practical insights for deploying vision-language models in resource-constrained or imbalanced settings.

\begin{figure}[!t]
\setlength{\abovecaptionskip}{0.2cm}
\includegraphics[width=3.5in]{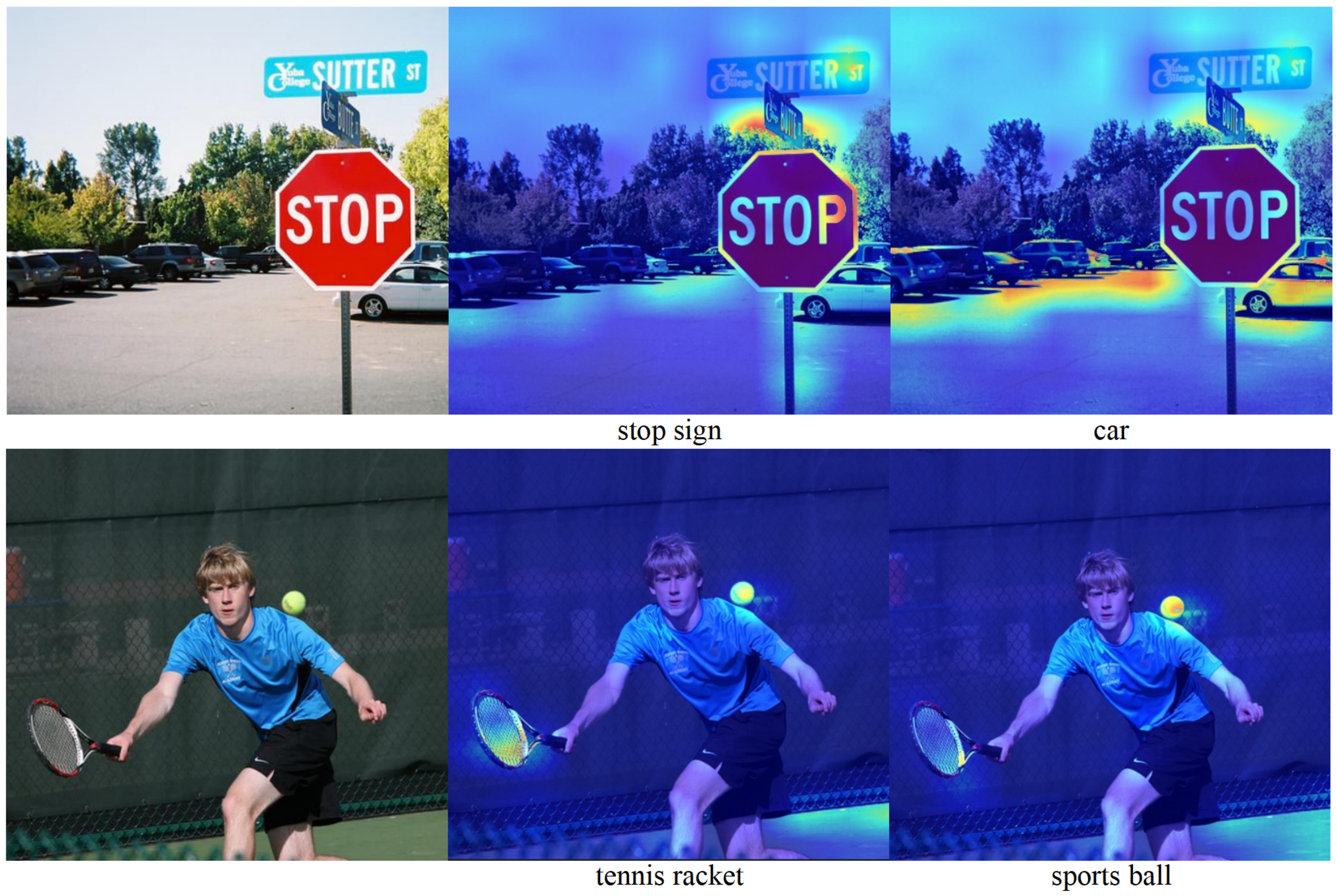}
\caption{Visualization of attention maps generated by {\ours}. The first column is the original image. Subsequent columns display attention maps queried by label-wise text features indicated below (``car'' is a head class, ``sports ball'' and ``tennis racket'' are medium classes, ``stop sign'' is a tail class).}
\label{fig:attention_map}
\end{figure}

\subsection{Further Analysis}
The effectiveness of our approach has been evaluated through comparisons with existing methods and detailed ablation studies. In this section, we conduct further experiments to offer deeper insights into the proposed method.

\subsubsection{\textbf{Attention Visualization}}
We present attention visualization generated by {\ours} in Fig.~\ref{fig:attention_map} to demonstrate the ability of our learned label features to capture semantic dependencies. The global image features from the image encoder pass through a projection layer, complicating direct extraction of local features from the vision transformer base backbone. To generate region-specific attention maps, we crop each image into $14 \times 14$ patches and compute cosine similarities between patch features and query label-wise text features.

The visualizations reveal that the refined features extract more discriminative and contextually rich information. For instance, attention maps for ``tennis racket'' highlight regions associated with ``sports ball'' due to their co-occurrence in sports scenes, while maps for ``stop sign'' emphasize areas linked to ``car'' in traffic contexts. This bidirectional semantic correlation enables mutual reinforcement among labels. Such mechanisms prove particularly beneficial for objects requiring extensive contextual cues, especially tail classes like ``stop sign'', which suffer from data scarcity. By leveraging correlations with frequent head classes such as ``car'' and medium classes like ``sports ball'' and ``tennis racket'', the model enhances recognition accuracy for underrepresented labels, mitigating long-tailed imbalances in multi-label scenarios.
These observations underscore how our model propagates knowledge across class frequencies, fostering robust feature representations that align visual and textual semantics effectively.

\begin{figure*}[!t]
\setlength{\abovecaptionskip}{0.cm} 
    \centering
  \begin{overpic}[width=16.cm]{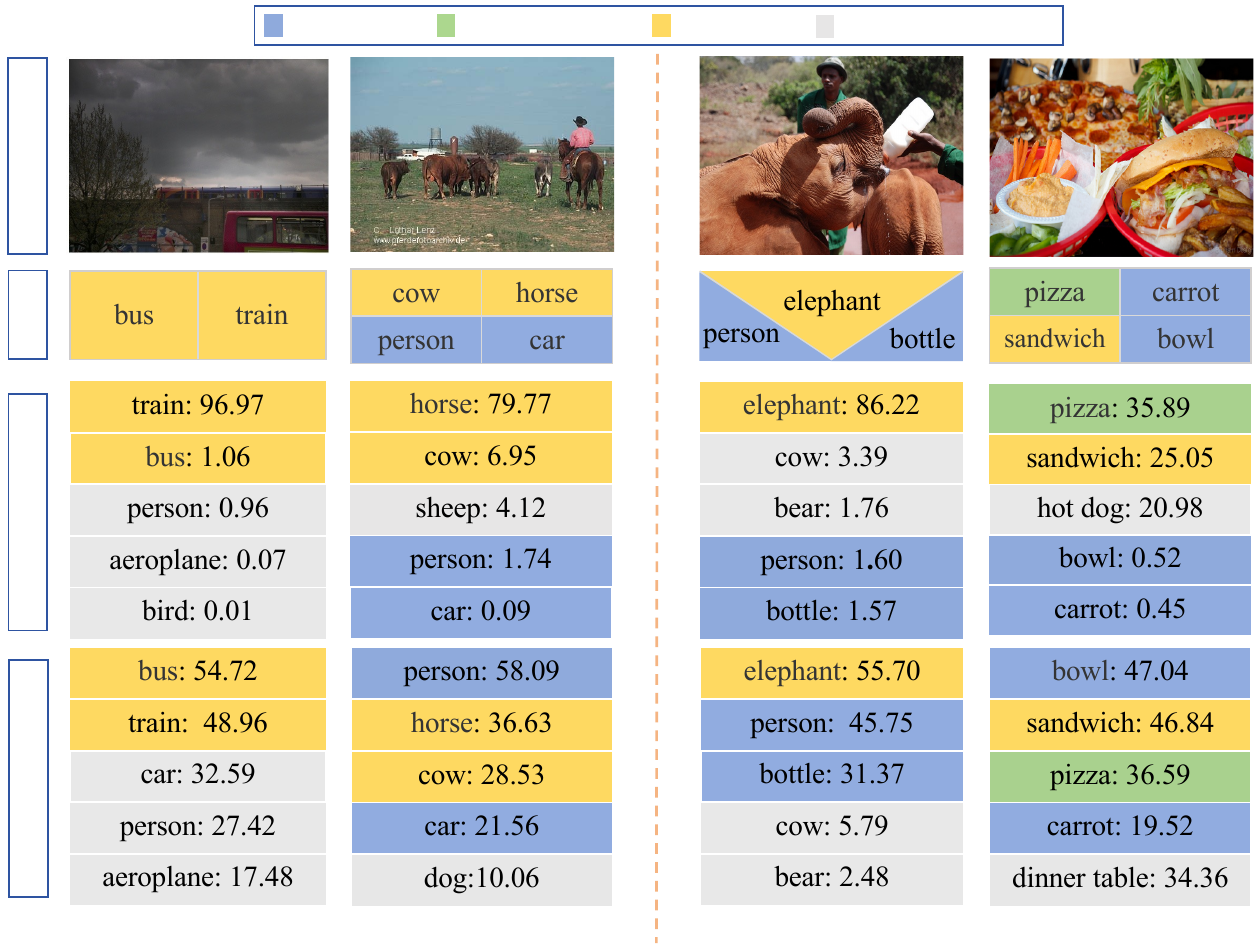}
        \put(230, 728){\small head classes}
        \put(368, 728){\small medium classes}
        \put(541, 728){\small tail classes}
        \put(671, 728){\small misclassified classes}
        \put(14, 578) {\rotatebox{90}{\small input images}} 
        \put(14, 480) {\rotatebox{90}{\small labels}} 
        \put(14, 265) {\rotatebox{90}{\small LMPT's predictions}} 
        \put(14, 44) {\rotatebox{90}{\small {\ours}'s predictions}} 
        \put(100, 10){\small (a) two samples from the VOC-LT dataset.}
        \put(600, 10){\small (b) two samples from the COCO-LT dataset.}
    \end{overpic}
    \caption{Comparison of the top-5 prediction (\%) visualizations between {\ours} and LMPT.}
 \label{fig:qualitative_pre}
\end{figure*}
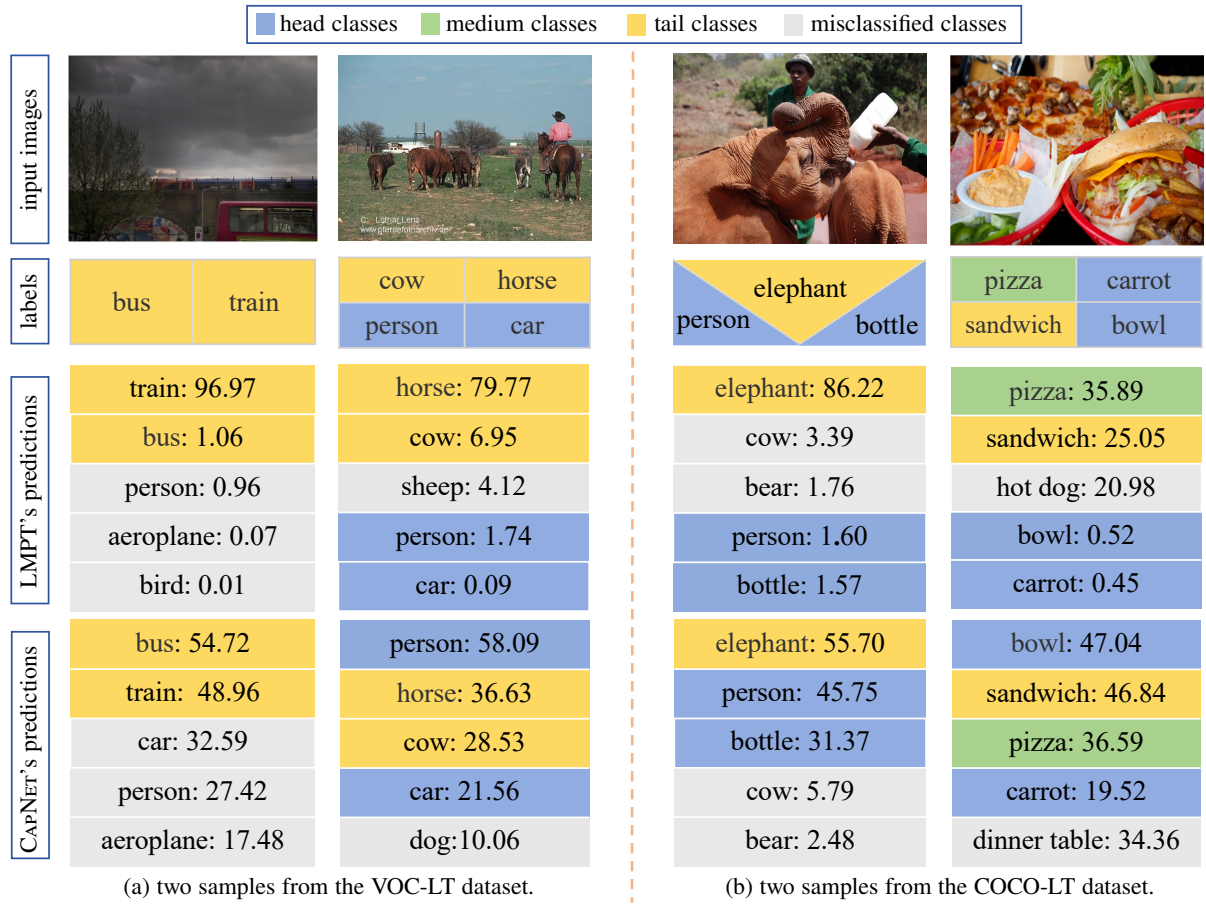

\subsubsection{\textbf{Qualitative Prediction Analysis}}
To highlight the superiority of our method over LMPT in managing long-tailed multi-label data, we perform a qualitative assessment on VOC-LT and COCO-LT. Fig.~\ref{fig:qualitative_pre} presents samples with ground truth labels and top-5 predictions, sorted by confidence, color-coded by class frequency: blue for head, green for medium, yellow for tail, and gray for misclassifications.

LMPT leverages image-caption supervision to capture inter-class relations but employs softmax, which skews confidence toward dominant labels. In multi-label cases, this suppresses secondary labels, e.g., in the first VOC-LT sample, train (head) at 96.97\% overshadows bus (head) at 1.06\%. Similarly, in the second, horse (medium) at 79.77\% diminishes cow (medium) at 6.95\% and person (head) at 1.74\%. Our sigmoid-based approach ensures equitable class treatment, producing balanced confidence. For the same samples, we assign 54.72\% to bus and 48.96\% to train; 58.09\% to person, 36.63\% to horse, and 28.53\% to cow, better capturing co-occurrences.
On COCO-LT, tail classes gain substantially: in the third sample, LMPT's elephant (tail) at 86.22\% relegates person (head) to 1.60\% and bottle (medium) to 1.57\%, while we elevate them to 45.75\% and 31.37\% beside elephant at 55.70\%. In the fourth, LMPT undervalues bowl (tail) at 0.52\% and carrot (tail) at 0.45\%; we boost them to 47.04\% and 19.52\%, with sandwich (medium) at 46.84\% and pizza (medium) at 36.59\%.
Graph-based label correspondences amplify related category confidences, aiding tail classes via knowledge transfer from frequent ones, thus overcoming softmax biases for robust multi-label recognition.

\subsubsection{\textbf{Performance on NUS-WIDE Dataset}}
NUS-WIDE, a real-world web image dataset, comprises 269648 Flickr images annotated with 81 visual concepts. We utilize the official train/test split, consisting of 161789 training images and 107859 test images. The dataset exhibits a pronounced long-tail distribution, exemplified by the head class ``sky'' with 44255 training samples and the tail class ``map'' with only 40 samples. Classes are categorized into three groups based on training sample counts: head classes exceed 8000 samples, medium classes range from 1000 to 8000 samples, and tail classes fall below 1000 samples. \looseness=-1

\begin{figure}[!t]
\setlength{\abovecaptionskip}{-0.2cm} 
    \centering
  \begin{overpic}[width=8.8cm]{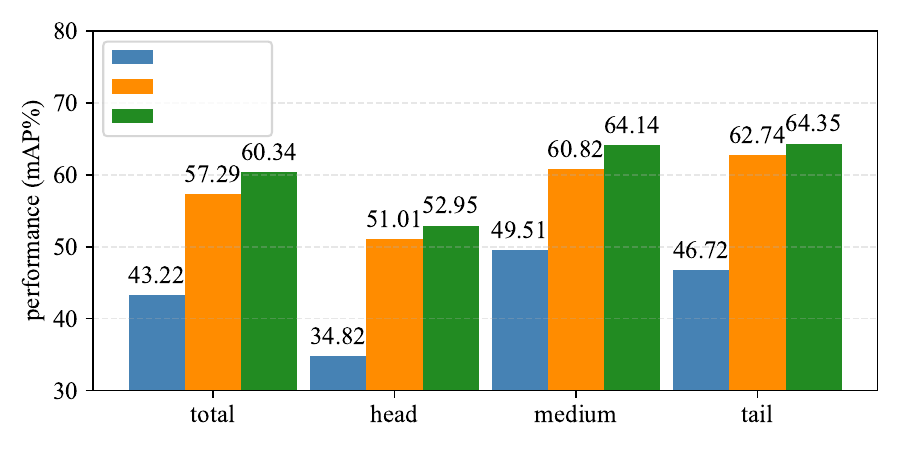}
        \put(188, 430){\tiny ZS-CLIP}
        \put(188, 395){\tiny FT-CLIP}
        \put(188, 362){\tiny {\ours}}
    \end{overpic}
    \caption{Performance (mAP\%) comparisons on NUS-WIDE dataset.}
 \label{fig:nus}
\end{figure}

Fig.~\ref{fig:nus} compares our method, {\ours}, with zero-shot CLIP (ZS-CLIP) and fine-tuning CLIP (FT-CLIP) on the NUS-WIDE dataset. 
FT-CLIP is incorporated with prompt tuning and AdaptFormer, supervised by Focal loss. As illustrated in Fig.~\ref{fig:nus}, the total mAP reaches 60.34\% with {\ours}, surpassing ZS-CLIP at 43.22\% and FT-CLIP at 57.29\%. This improvement stems from fine-tuning with a sigmoid layer and leveraging label correlations, enhancing multi-label recognition across long-tailed distributions.
Breaking down performance by class partition, head classes show {\ours} at 60.82\%, exceeding ZS-CLIP at 34.82\% and FT-CLIP at 51.01\%, reflecting robust handling of frequent categories. Medium classes achieve 64.14\% with {\ours}, compared to 49.51\% for ZS-CLIP and 52.95\% for FT-CLIP, indicating effective adaptation to moderately represented classes. Tail classes benefit most, with {\ours} reaching 64.35\%, significantly ahead of ZS-CLIP at 46.72\% and FT-CLIP at 62.74\%, demonstrating the method's strength in addressing underrepresented labels via correlation-based feature propagation. The results show that FT-CLIP for multi-label scenarios substantially boosts mAP, particularly for tail classes. Unlike recent approaches that preserve CLIP's original structure to mitigate modality gaps, our re-adjustment of CLIP's output and realignment of encoders proves more effective for long-tailed multi-label tasks, offering a balanced enhancement across all class frequencies.

\section{Conclusion}
\label{sec:con} 
In this paper, we introduced the Correlation Adaptation Prompt Network ({\ours}), a novel end-to-end framework designed to harness the power of pre-trained vision-language models (VLMs) like CLIP for addressing the challenges of long-tailed multi-label visual recognition {\ours} models label correlations using a graph convolutional network derived from CLIP's textual encoder and a distribution-balanced Focal loss. Furthermore, test-time ensembling enhances generalization by aggregating diverse predictions, ensuring robustness across varied backbones such as ViT and ResNet. The parameter-efficient fine-tuning via AdaptFormer effectively prevents overfitting on tail classes.
Extensive experiments on benchmarks, including VOC-LT, COCO-LT, and NUS-WIDE, demonstrated {\ours}'s superiority over state-of-the-art methods. On VOC-LT, {\ours} achieved a mean average precision (mAP) of 93.03\%, outperforming baselines like LMPT by 5.15\% overall. Similarly, on COCO-LT, it attained 76.36\% mAP, surpassing prior approaches by up to 10.17\%, particularly in medium and tail categories. These results validate {\ours}'s ability to mitigate long-tailed imbalances while maintaining high performance on head classes, underscoring its practical utility for real-world applications.
Future work could explore extending {\ours} to open-vocabulary settings or integrating it with emerging multimodal architectures to further enhance scalability and adaptability in dynamic environments.

\bibliographystyle{IEEEtran}
\bibliography{ref}

\begin{thebibliography}{10}
\providecommand{\url}[1]{#1}
\csname url@samestyle\endcsname
\providecommand{\newblock}{\relax}
\providecommand{\bibinfo}[2]{#2}
\providecommand{\BIBentrySTDinterwordspacing}{\spaceskip=0pt\relax}
\providecommand{\BIBentryALTinterwordstretchfactor}{4}
\providecommand{\BIBentryALTinterwordspacing}{\spaceskip=\fontdimen2\font plus
\BIBentryALTinterwordstretchfactor\fontdimen3\font minus
  \fontdimen4\font\relax}
\providecommand{\BIBforeignlanguage}[2]{{%
\expandafter\ifx\csname l@#1\endcsname\relax
\typeout{** WARNING: IEEEtran.bst: No hyphenation pattern has been}%
\typeout{** loaded for the language `#1'. Using the pattern for}%
\typeout{** the default language instead.}%
\else
\language=\csname l@#1\endcsname
\fi
#2}}
\providecommand{\BIBdecl}{\relax}
\BIBdecl

\bibitem{ZhangKHYF23}
Y.~Zhang, B.~Kang, B.~Hooi, S.~Yan, and J.~Feng, ``Deep long-tailed learning:
  {A} survey,'' \emph{{IEEE} Transactions on Pattern Analysis and Machine
  Intelligence}, vol.~45, no.~9, pp. 10\,795--10\,816, 2023.

\bibitem{yun2019cutmix}
S.~Yun, D.~Han, S.~J. Oh, S.~Chun, J.~Choe, and Y.~Yoo, ``Cut{M}ix:
  {R}egularization strategy to train strong classifiers with localizable
  features,'' in \emph{Proceedings of the {IEEE/CVF} International Conference
  on Computer Vision, Seoul, South Korea}, 2019, pp. 6023--6032.

\bibitem{alshammari2022long}
S.~Alshammari, Y.-X. Wang, D.~Ramanan, and S.~Kong, ``Long-tailed recognition
  via weight balancing,'' in \emph{Proceedings of the IEEE/CVF Conference on
  Computer Vision and Pattern Recognition, New Orleans, LA, USA}, 2022, pp.
  6897--6907.

\bibitem{menon2021long}
A.~K. Menon, S.~Jayasumana, A.~S. Rawat, H.~Jain, A.~Veit, and S.~Kumar,
  ``Long-tail learning via logit adjustment,'' in \emph{Proceedings of the 9th
  International Conference on Learning Representations, Virtual Event}, 2021,
  pp. 1--24.

\bibitem{wu2020distribution}
T.~Wu, Q.~Huang, Z.~Liu, Y.~Wang, and D.~Lin, ``Distribution-balanced loss for
  multi-label classification in long-tailed datasets,'' in \emph{Proceedings of
  the 16th European Conference on Computer Vision, Glasgow, UK}.\hskip 1em plus
  0.5em minus 0.4em\relax Springer, 2020, pp. 162--178.

\bibitem{guo2021long}
H.~Guo and S.~Wang, ``Long-tailed multi-label visual recognition by
  collaborative training on uniform and re-balanced samplings,'' in
  \emph{Proceedings of the IEEE/CVF Conference on Computer Vision and Pattern
  Recognition}, 2021, pp. 15\,089--15\,098.

\bibitem{chen2023class}
J.~Chen and S.~Li, ``Class-aware learning for imbalanced multi-label
  classification,'' in \emph{Proceedings of the {IEEE} 5th International
  Conference on Civil Aviation Safety and Information Technology}, 2023, pp.
  903--907.

\bibitem{xia2023lmpt}
P.~Xia, D.~Xu, M.~Hu, L.~Ju, and Z.~Ge, ``{LMPT}: Prompt tuning with
  class-specific embedding loss for long-tailed multi-label visual
  recognition,'' in \emph{Proceedings of the 3rd Workshop on Advances in
  Language and Vision Research}, 2024, pp. 26--36.

\bibitem{Yan0MH024}
J.~Yan, S.~Huang, N.~Mu, L.~Huangfu, and B.~Liu, ``Category-prompt refined
  feature learning for long-tailed multi-label image classification,'' in
  \emph{Proceedings of the 32nd {ACM} International Conference on Multimedia,
  Melbourne, VIC, Australia}, 2024, pp. 2146--2155.

\bibitem{TimmermannJKL25}
C.~Timmermann, S.~Jung, M.~Kim, and W.~Lee, ``{LM-CLIP:} {A}dapting positive
  asymmetric loss for long-tailed multi-label classification,'' \emph{{IEEE}
  Access}, vol.~13, pp. 71\,053--71\,065, 2025.

\bibitem{radford2021learning}
A.~Radford, J.~W. Kim, C.~Hallacy, A.~Ramesh, G.~Goh, S.~Agarwal, G.~Sastry,
  A.~Askell, P.~Mishkin, J.~Clark, G.~Krueger, and I.~Sutskever, ``Learning
  transferable visual models from natural language supervision,'' in
  \emph{Proceedings of the 38th International Conference on Machine Learning,
  Virtual Event}, vol. 139, 2021, pp. 8748--8763.

\bibitem{chen2019multi}
Z.~Chen, X.~Wei, P.~Wang, and Y.~Guo, ``Multi-label image recognition with
  graph convolutional networks,'' in \emph{Proceedings of the {IEEE/CVF}
  Conference on Computer Vision and Pattern Recognition, Long Beach, CA, USA},
  2019, pp. 5177--5186.

\bibitem{ding2023exploring}
Z.~Ding, A.~Wang, H.~Chen, Q.~Zhang, P.~Liu, Y.~Bao, W.~Yan, and J.~Han,
  ``Exploring structured semantic prior for multi-label recognition with
  incomplete labels,'' in \emph{Proceedings of the {IEEE/CVF} Conference on
  Computer Vision and Pattern Recognition, Vancouver, BC, Canada}, 2023, pp.
  3398--3407.

\bibitem{shen2016relay}
L.~Shen, Z.~Lin, and Q.~Huang, ``Relay backpropagation for effective learning
  of deep convolutional neural networks,'' in \emph{Proceedings of the 14th
  European Conference on Computer Vision, Amsterdam, The Netherlands}, vol.
  9911, 2016, pp. 467--482.

\bibitem{everingham2015pascal}
M.~Everingham, S.~Eslami, L.~Van~Gool, C.~K. Williams, J.~Winn, and
  A.~Zisserman, ``The pascal visual object classes challenge: A
  retrospective,'' \emph{International Journal of Computer Vision}, vol. 111,
  no.~1, pp. 98--136, 2015.

\bibitem{lin2014microsoft}
T.-Y. Lin, M.~Maire, S.~Belongie, J.~Hays, P.~Perona, D.~Ramanan,
  P.~Doll{\'a}r, and C.~L. Zitnick, ``Microsoft {COCO:} {C}ommon objects in
  context,'' in \emph{Proceedings of the 13th European Conference on Computer
  Vision, Zurich, Switzerland}, 2014, pp. 740--755.

\bibitem{zhang2025la}
X.~Zhang, R.~He, C.~Jiao, D.~Fang, M.~Li, Z.~Zeng, C.~Chen, and H.~Zhuang,
  ``{L3A}: {L}abel-augmented analytic adaptation for multi-label class
  incremental learning,'' in \emph{Proceedings of the 42nd International
  Conference on Machine Learning, Vancouver, Canada}, 2025, pp. 1--14.

\bibitem{wen2025learning}
J.~Wen, Y.~Liu, Z.~Tang, Y.~He, Y.~Chen, M.~Li, and C.~Liu, ``Learning compact
  semantic information for incomplete multi-view missing multi-label
  classification,'' in \emph{Proceedings of the 42nd International Conference
  on Machine Learning, Vancouver, Canada}, 2025, pp. 1--12.

\bibitem{zhang2013review}
M.-L. Zhang and Z.-H. Zhou, ``A review on multi-label learning algorithms,''
  \emph{IEEE Transactions on Knowledge and Data Engineering}, vol.~26, no.~8,
  pp. 1819--1837, 2013.

\bibitem{tsoumakas2007multi}
G.~Tsoumakas and I.~Katakis, ``Multi-label classification: {A}n overview,''
  \emph{International Journal of Data Warehousing and Mining}, vol.~3, no.~3,
  pp. 1--13, 2007.

\bibitem{zhang2007ml}
M.-L. Zhang and Z.-H. Zhou, ``{ML-KNN}: {A} lazy learning approach to
  multi-label learning,'' \emph{Pattern recognition}, vol.~40, no.~7, pp.
  2038--2048, 2007.

\bibitem{wang2016cnn}
J.~Wang, Y.~Yang, J.~Mao, Z.~Huang, C.~Huang, and W.~Xu, ``{CNN-RNN:} {A}
  unified framework for multi-label image classification,'' in
  \emph{Proceedings of the {IEEE} Conference on Computer Vision and Pattern
  Recognition, Las Vegas, NV, USA}, 2016, pp. 2285--2294.

\bibitem{chen2019learning}
T.~Chen, M.~Xu, X.~Hui, H.~Wu, and L.~Lin, ``Learning semantic-specific graph
  representation for multi-label image recognition,'' in \emph{Proceedings of
  the {IEEE/CVF} International Conference on Computer Vision, Seoul, South
  Korea}, 2019, pp. 522--531.

\bibitem{wang2020multi}
Y.~Wang, D.~He, F.~Li, X.~Long, Z.~Zhou, J.~Ma, and S.~Wen, ``Multi-label
  classification with label graph superimposing,'' in \emph{Proceedings of the
  34th {AAAI} Conference on Artificial Intelligence, New York, NY, USA}, 2020,
  pp. 12\,265--12\,272.

\bibitem{ye2020attention}
J.~Ye, J.~He, X.~Peng, W.~Wu, and Y.~Qiao, ``Attention-driven dynamic graph
  convolutional network for multi-label image recognition,'' in
  \emph{Proceedings of the 16th European Conference on Computer Vision,
  Glasgow, UK}, 2020, pp. 649--665.

\bibitem{huynh2020interactive}
D.~Huynh and E.~Elhamifar, ``Interactive multi-label {CNN} learning with
  partial labels,'' in \emph{Proceedings of the {IEEE/CVF} Conference on
  Computer Vision and Pattern Recognition, Seattle, WA, USA}, 2020, pp.
  9423--9432.

\bibitem{chen2022structured}
T.~Chen, T.~Pu, H.~Wu, Y.~Xie, and L.~Lin, ``Structured semantic transfer for
  multi-label recognition with partial labels,'' in \emph{Proceedings of the
  36th {AAAI} Conference on Artificial Intelligence, Virtual Event}, 2022, pp.
  339--346.

\bibitem{sun2022dualcoop}
X.~Sun, P.~Hu, and K.~Saenko, ``Dual{C}o{O}p: {F}ast adaptation to multi-label
  recognition with limited annotations,'' in \emph{Advances in Neural
  Information Processing Systems 35, New Orleans, LA, USA}, 2022.

\bibitem{guo2023texts}
Z.~Guo, B.~Dong, Z.~Ji, J.~Bai, Y.~Guo, and W.~Zuo, ``Texts as images in prompt
  tuning for multi-label image recognition,'' in \emph{Proceedings of the
  {IEEE/CVF} Conference on Computer Vision and Pattern Recognition, Vancouver,
  BC, Canada}, 2023, pp. 2808--2817.

\bibitem{fang2023revisiting}
C.~Fang, D.~Zhang, W.~Zheng, X.~Li, L.~Yang, L.~Cheng, and J.~Han, ``Revisiting
  long-tailed image classification: {S}urvey and benchmarks with new evaluation
  metrics,'' \emph{CoRR}, vol. abs/2302.01507, 2023.

\bibitem{he2025partialclip}
K.~He, W.~Tang, T.~Wei, and M.~Zhang, ``Tuning the right foundation models is
  what you need for partial label learning,'' \emph{CoRR}, vol. abs/2506.05027,
  2025.

\bibitem{WeiL20}
T.~Wei and Y.~Li, ``Does tail label help for large-scale multi-label
  learning?'' \emph{{IEEE} Transactions on Neural Networks and Learning
  Systems}, vol.~31, no.~7, pp. 2315--2324, 2020.

\bibitem{chawla2002smote}
N.~V. Chawla, K.~W. Bowyer, L.~O. Hall, and W.~P. Kegelmeyer, ``{SMOTE:}
  synthetic minority over-sampling technique,'' \emph{Journal of Artificial
  Intelligence Research}, vol.~16, pp. 321--357, 2002.

\bibitem{chris2003class}
D.~Chris and C.~Robert, ``Class imbalance and cost sensitivity: Why
  undersampling beats oversampling,'' in \emph{ICML-KDD 2003 Workshop: Learning
  from Imbalanced Datasets}, vol.~3, 2003, pp. 1--8.

\bibitem{tang2023demipl}
W.~Tang, W.~Zhang, and M.-L. Zhang, ``Disambiguated attention embedding for
  multi-instance partial-label learning,'' in \emph{Advances in Neural
  Information Processing Systems 36, New Orleans, LA, USA}, 2023, pp.
  56\,756--56\,771.

\bibitem{kang2019decoupling}
B.~Kang, S.~Xie, M.~Rohrbach, Z.~Yan, A.~Gordo, J.~Feng, and Y.~Kalantidis,
  ``Decoupling representation and classifier for long-tailed recognition,'' in
  \emph{Proceedings of the 8th International Conference on Learning
  Representations, Addis Ababa, Ethiopia}, 2020, pp. 1--16.

\bibitem{cui2021parametric}
J.~Cui, Z.~Zhong, S.~Liu, B.~Yu, and J.~Jia, ``Parametric contrastive
  learning,'' in \emph{Proceedings of the {IEEE/CVF} International Conference
  on Computer Vision, Montreal, QC, Canada}, 2021, pp. 695--704.

\bibitem{li2022targeted}
T.~Li, P.~Cao, Y.~Yuan, L.~Fan, Y.~Yang, R.~S. Feris, P.~Indyk, and D.~Katabi,
  ``Targeted supervised contrastive learning for long-tailed recognition,'' in
  \emph{Proceedings of the IEEE/CVF Conference on Computer Vision and Pattern
  Recognition}, 2022, pp. 6918--6928.

\bibitem{tang2023miplgp}
W.~Tang, W.~Zhang, and M.-L. Zhang, ``Multi-instance partial-label learning:
  {T}owards exploiting dual inexact supervision,'' \emph{Science China
  Information Sciences}, vol.~67, no.~3, pp. 132\,103:1--132\,103:14, 2024.

\bibitem{lin2020focal}
T.~Lin, P.~Goyal, R.~B. Girshick, K.~He, and P.~Doll{\'{a}}r, ``Focal loss for
  dense object detection,'' \emph{{IEEE} Transactions on Pattern Analysis and
  Machine Intelligence}, no.~2, pp. 318--327, 2020.

\bibitem{park2021influence}
S.~Park, J.~Lim, Y.~Jeon, and J.~Y. Choi, ``Influence-balanced loss for
  imbalanced visual classification,'' in \emph{Proceedings of the {IEEE/CVF}
  International Conference on Computer Vision, Montreal, QC, Canada}, 2021, pp.
  715--724.

\bibitem{cao2019learning}
K.~Cao, C.~Wei, A.~Gaidon, N.~Arechiga, and T.~Ma, ``Learning imbalanced
  datasets with label-distribution-aware margin loss,'' \emph{Advances in
  neural information processing systems}, vol.~32, 2019.

\bibitem{cai2021ace}
J.~Cai, Y.~Wang, and J.~Hwang, ``{ACE:} ally complementary experts for solving
  long-tailed recognition in one-shot,'' in \emph{Proceedings of the {IEEE/CVF}
  International Conference on Computer Vision, Montreal, QC, Canada}, 2021, pp.
  112--121.

\bibitem{wang2020long}
X.~Wang, L.~Lian, Z.~Miao, Z.~Liu, and S.~X. Yu, ``Long-tailed recognition by
  routing diverse distribution-aware experts,'' in \emph{Proceedings of the 9th
  International Conference on Learning Representations, Virtual Event}, 2021,
  pp. 1--15.

\bibitem{ma2021simple}
T.~Ma, S.~Geng, M.~Wang, J.~Shao, J.~Lu, H.~Li, P.~Gao, and Y.~Qiao, ``A simple
  long-tailed recognition baseline via vision-language model,'' \emph{CoRR},
  vol. abs/2111.14745, 2021.

\bibitem{tian2022vl}
C.~Tian, W.~Wang, X.~Zhu, J.~Dai, and Y.~Qiao, ``{VL-LTR:} learning class-wise
  visual-linguistic representation for long-tailed visual recognition,'' in
  \emph{Proceedings of the 17th European Conference on Computer Vision, Tel
  Aviv, Israel}, vol. 13685, 2022, pp. 73--91.

\bibitem{zhou2020bbn}
B.~Zhou, Q.~Cui, X.~Wei, and Z.~Chen, ``{BBN:} {B}ilateral-branch network with
  cumulative learning for long-tailed visual recognition,'' in
  \emph{Proceedings of the {IEEE/CVF} Conference on Computer Vision and Pattern
  Recognition, Seattle, WA, USA}, 2020, pp. 9716--9725.

\bibitem{pennington2014glove}
J.~Pennington, R.~Socher, and C.~D. Manning, ``Glo{V}e: {G}lobal vectors for
  word representation,'' in \emph{Proceedings of the Conference on Empirical
  Methods in Natural Language Processing, Doha, Qatar}, 2014, pp. 1532--1543.

\bibitem{ZhangLZOTZ23}
W.~Zhang, C.~Liu, L.~Zeng, B.~C. Ooi, S.~Tang, and Y.~Zhuang, ``Learning in
  imperfect environment: Multi-label classification with long-tailed
  distribution and partial labels,'' in \emph{Proceedings of the {IEEE/CVF}
  International Conference on Computer Vision, Paris, France}, 2023, pp.
  1423--1432.

\bibitem{LinPCXQC24}
D.~Lin, T.~Peng, R.~Chen, X.~Xie, X.~Qin, and Z.~Cui, ``Distributionally robust
  loss for long-tailed multi-label image classification,'' in \emph{Proceedings
  of the 18th European Conference on Computer Vision, Milan, Italy}, vol.
  15091, 2024, pp. 417--433.

\bibitem{TaoLWZCLHC25}
Z.~Tao, S.~Li, W.~Wan, J.~Zheng, J.~Chen, Y.~Li, S.~Huang, and S.~Chen,
  ``{MLC-NC:} {L}ong-tailed multi-label image classification through the lens
  of neural collapse,'' in \emph{Proceedings of the 39th AAAI Conference on
  Artificial Intelligence, Philadelphia, PA, {USA}}, 2025, pp.
  20\,850--20\,858.

\bibitem{shi2023parameter}
J.~Shi, T.~Wei, Z.~Zhou, J.~Shao, X.~Han, and Y.~Li, ``Long-tail learning with
  foundation model: {H}eavy fine-tuning hurts,'' in \emph{Proceedings of the
  41st International Conference on Machine Learning, Vienna, Austria}, 2024.

\bibitem{shi2025lift}
J.~Shi, T.~Wei, and Y.~Li, ``{LIFT+:} {L}ightweight fine-tuning for long-tail
  learning,'' \emph{CoRR}, vol. abs/2504.13282, 2025.

\bibitem{zhangfx2025peft}
D.~Zhang, T.~Feng, L.~Xue, Y.~Wang, Y.~Dong, and J.~Tang, ``Parameter-efficient
  fine-tuning for foundation models,'' \emph{CoRR}, vol. abs/2501.13787, 2025.

\bibitem{zhou2022conditional}
K.~Zhou, J.~Yang, C.~C. Loy, and Z.~Liu, ``Conditional prompt learning for
  vision-language models,'' in \emph{Proceedings of the 33rd IEEE/CVF
  Conference on Computer Vision and Pattern Recognition, New Orleans, LA, USA},
  2022, pp. 16\,795--16\,804.

\bibitem{dosovitskiy2021image}
A.~Dosovitskiy, L.~Beyer, A.~Kolesnikov, D.~Weissenborn, X.~Zhai,
  T.~Unterthiner, M.~Dehghani, M.~Minderer, G.~Heigold, S.~Gelly, J.~Uszkoreit,
  and N.~Houlsby, ``An image is worth 16x16 words: Transformers for image
  recognition at scale,'' in \emph{Proceedings of the 9th International
  Conference on Learning Representations, Virtual Event}, 2021, pp. 1--21.

\bibitem{jia2022visual}
M.~Jia, L.~Tang, B.-C. Chen, C.~Cardie, S.~Belongie, B.~Hariharan, and S.-N.
  Lim, ``Visual prompt tuning,'' in \emph{Proceedings of the 17th European
  Conference on Computer Vision, Tel Aviv, Israel}, 2022, pp. 709--727.

\bibitem{chen2022adaptformer}
S.~Chen, C.~GE, Z.~Tong, J.~Wang, Y.~Song, J.~Wang, and P.~Luo, ``Adaptformer:
  Adapting vision transformers for scalable visual recognition,'' in
  \emph{Advances in Neural Information Processing Systems 35}, 2022, pp.
  16\,664--16\,678.

\bibitem{he2016deep}
K.~He, X.~Zhang, S.~Ren, and J.~Sun, ``Deep residual learning for image
  recognition,'' in \emph{Proceedings of the 29th {IEEE/CVF} Conference on
  Computer Vision and Pattern Recognition, Las Vegas, NV, USA}, 2016, pp.
  770--778.

\bibitem{VaswaniSPUJGKP17}
A.~Vaswani, N.~Shazeer, N.~Parmar, J.~Uszkoreit, L.~Jones, A.~N. Gomez,
  L.~Kaiser, and I.~Polosukhin, ``Attention is all you need,'' in
  \emph{Advances in Neural Information Processing Systems 30, Long Beach, CA,
  {USA}}, 2017, pp. 5998--6008.

\bibitem{RidnikBZNFPZ21}
T.~Ridnik, E.~B. Baruch, N.~Zamir, A.~Noy, I.~Friedman, M.~Protter, and
  L.~Zelnik{-}Manor, ``Asymmetric loss for multi-label classification,'' in
  \emph{Proceedings of the {IEEE/CVF} International Conference on Computer
  Vision, Montreal, QC, Canada}, 2021, pp. 82--91.

\bibitem{liu2019large}
Z.~Liu, Z.~Miao, X.~Zhan, J.~Wang, B.~Gong, and S.~X. Yu, ``Large-scale
  long-tailed recognition in an open world,'' in \emph{Proceedings of the
  {IEEE/CVF} Conference on Computer Vision and Pattern Recognition, Long Beach,
  CA, USA}, 2019, pp. 2537--2546.

\bibitem{cui2019class}
Y.~Cui, M.~Jia, T.-Y. Lin, Y.~Song, and S.~Belongie, ``Class-balanced loss
  based on effective number of samples,'' in \emph{Proceedings of the
  {IEEE/CVF} Conference on Computer Vision and Pattern Recognition, Long Beach,
  CA, USA}, 2019, pp. 9268--9277.

\bibitem{zhou2022learning}
K.~Zhou, J.~Yang, C.~C. Loy, and Z.~Liu, ``Learning to prompt for
  vision-language models,'' \emph{International Journal of Computer Vision},
  vol. 130, no.~9, pp. 2337--2348, 2022.

\end{thebibliography}

\ifCLASSOPTIONcaptionsoff
  \newpage
\fi

\end{document}